%% file: main.tex
\documentclass{article}

\usepackage{arxiv}
\usepackage{url}            
\usepackage{booktabs}       
\usepackage{amsfonts}       
\usepackage{nicefrac}       
\usepackage{lipsum}
\usepackage[utf8]{inputenc}      
\usepackage[T1]{fontenc}         
\usepackage{amsmath, amssymb}    
\usepackage{graphicx}            
\usepackage[hidelinks]{hyperref}            
\usepackage{geometry}            
\usepackage{enumitem}            
\usepackage[table]{xcolor}
\usepackage{microtype}           
\usepackage{threeparttable}
\usepackage{multirow}
\usepackage{pdflscape} 
\usepackage{caption}
\usepackage{subcaption}
\usepackage{float}
\usepackage{subfloat}
\usepackage{tabularx}
\usepackage{array}

\input{macros}
\newcommand{\mcomma}{,\allowbreak\ }

\graphicspath{ {./figures/} }

\title{Domain Generalization for Smartphone-Based Human Activity Recognition: A Systematic Analysis of Components and Interactions}

\author{
 Otávio Oliveira Napoli \\
  Institute of Computing\\
  University of Campinas\\
  Av. Albert Einstein \\
  \texttt{otavio.napoli@ic.unicamp.br} \\
   \And
 Edson Borin \\
  Institute of Computing\\
  University of Campinas\\
  Av. Albert Einstein \\
  \texttt{borin@unicamp.br} \\
}

\begin{document}
\maketitle


\begin{abstract}
Smartphone-based Human Activity Recognition (HAR) models often degrade under distribution shifts caused by changes in users, devices, sensor placements, environments, and acquisition protocols. Domain Generalization (DG) addresses this problem by learning from source domains without access to target-domain data. Existing DG approaches span training objectives, representation initialization strategies, and architectural modifications. However, works commonly evaluate these components in isolation, even though they operate at different stages of the learning pipeline and can interact with one another and with the underlying model architecture.

We present a large-scale controlled benchmark of DG for smartphone-based HAR, comprising more than $410{,}000$ experiments across four model architectures, thirteen training objectives including Empirical Risk Minimization (ERM), five initialization strategies, four architectural configurations, and two complementary distribution-shift scenarios: cross-dataset and cross-position. We evaluate each setting through systematic hyperparameter search and source-only model selection.

Our results show that individual DG components provide limited and highly conditional gains. 
Alternative objectives rarely outperform ERM consistently, self-supervised initialization is beneficial in specific settings, and architectural modifications (particularly Dynamic Domain Generalization) provide the clearest standalone improvements. 
Joint configurations, however, frequently outperform their individual components and exhibit complementary and, in several cases, super-additive interactions, although these effects remain strongly model- and shift-dependent. 
Class-level analysis further shows that the strongest configurations improve generalization by correcting a limited set of difficult, shift-sensitive decision boundaries, especially among locomotion and stair-related activities.

Finally, oracle checkpoint analysis reveals substantial unrealized performance: source-validation selection recovers only $53\%$ and $26\%$ of the available oracle gain in cross-dataset and cross-position settings, respectively. Overall, our findings indicate that progress in HAR domain generalization depends on jointly designing DG components and a robust model selection strategy.
\end{abstract}

\section{Introduction}
\label{sec:introduction}

Generalization is a central objective in machine learning (ML), referring to a model's ability to maintain reliable performance under unseen data and distribution shifts~\cite{vapnik1998statistical}. 
This challenge is particularly important in Human Activity Recognition (HAR), where variations in users, devices, sensor placements, environments, and acquisition protocols substantially alter inertial signals. 
Smartphones offer an attractive sensing platform due to their ubiquitous accelerometers and gyroscopes. 
However, models trained under controlled conditions often degrade considerably when deployed in new environments~\cite{napoli2024benchmark,lu2025harood}.

Most HAR studies assume training and test samples are independently and identically distributed (\iid), typically evaluating models through random splits or within-dataset cross-validation~\cite{ha2015multi,ha2016convolutional,yue2022ts2vec,shavit2021boosting,mekruksavanich2022deep,da2026benchmarking}. 
While appropriate for measuring in-domain recognition, these protocols provide limited evidence of robustness under realistic deployment, where target distributions differ from training data. 

Domain Generalization (DG) addresses this challenge by learning from multiple source domains and evaluating on an unseen target domain with a different distribution~\cite{wang2022generalizing}.
In HAR, this includes cross-subject, cross-position, and cross-device generalization, providing a more realistic assessment of deployment robustness.

A wide range of DG approaches has been proposed~\cite{wang2022generalizing,sagawa2019distributionally,krueger2021out,sun2016deep,huang2020self} including many specifically designed for time-series data and HAR~\cite{zhang2022self,lu2022domain,lu2022semantic,hong2024crosshar}. 
However, recent evidence suggests that generalization depends not only on the DG objective itself, but also on the interaction between representation initialization, backbone architecture, and optimization procedure~\cite{teterwak2025erm++,teterwak2025large,napoli2025domain,napoli2026components}. 
Existing HAR works primarily compare objective-based DG methods~\cite{napoli2024benchmark,lu2025harood,hong2024crosshar,napoli2025domain}, largely overlooking the influence of representation initialization and architectural modifications, which may lead to suboptimal conclusions about the effectiveness of DG methods and model architectures.

Thus, rather than studying DG methods solely through their learning objectives, we organize them according to the component of the learning pipeline they modify. Specifically, we consider three complementary dimensions: 
(i) representation initialization, which aims to improve feature quality before supervised DG training through strategies such as self-supervised pretraining; 
(ii) architectural modifications, which introduce specialized operations into the model to augment feature distributions or promote representations that are more robust to domain shifts; and 
(iii) objective-based methods, which modify the default supervised optimization through additional losses, regularization terms, or robustness constraints, while preserving the underlying model architecture.

Although these dimensions operate at different stages of learning, their effects are not necessarily independent. 
A component that provides little benefit in isolation may become effective when paired with a compatible backbone or architectural modification (as well as competitive, redundant, or detrimental). 
We therefore, based on recent evidences~\cite{teterwak2025erm++,teterwak2025large,napoli2025domain,napoli2026components}, hypothesize that DG performance is determined by the complete learning pipeline, including initialization, architecture, objective, and model-selection procedure, rather than by the DG objective alone.

To investigate this hypothesis, we conduct, to the best of our knowledge, the largest controlled DG benchmark for smartphone-based HAR to date, comprising more than $410{,}000$ experiments. 
The benchmark includes four representative HAR backbones (CNN-PFF, ResNet-SE-5, TS2Vec, and IMU-Transformer), thirteen training objectives including ERM, five representation initialization strategies, and four architectural modification modules. 
We evaluate these components under six cross-dataset (CD) and four cross-position (CP) leave-one-domain-out folds.

The main contributions of this work are summarized as follows:
\begin{itemize}
    \item We present a large-scale controlled benchmark of DG for smartphone-based HAR, comprising more than $410{,}000$ experiments across four model architectures, thirteen training objectives, five initialization strategies, four architectural configurations, and cross-dataset and cross-position shifts.

    \item We introduce a component-oriented full-factorial evaluation that separates representation initialization, architectural modifications, and objective-based DG, enabling controlled analysis of their individual and joint effects.

    \item We show that component compatibility strongly determines DG performance. Individual interventions provide limited, highly conditional gains, whereas joint configurations can exhibit complementary or super-additive effects, depending on the model and distribution shift. Architectural modifications, particularly DDG, provide the clearest standalone gains.
    
    \item We show that some of the largest class-level gains of winning DG configurations are concentrated on difficult, shift-sensitive decision boundaries, particularly those involving walking, running, and stair activities, providing insight into where the observed generalization improvements arise.

    \item We identify source-only model selection as a major bottleneck. Oracle checkpoint analysis shows that source-validation selection recovers only $53\%$ of the available oracle gain in cross-dataset and $26\%$ in cross-position settings at the highest compositional level.
\end{itemize}

The remainder of this paper is organized as follows.
Section~\ref{sec:dg-problem} formalizes the DG problem, while Section~\ref{sec:har-generalization-challenges} introduces the distribution shifts considered in smartphone-based HAR.
Section~\ref{sec:dg-methods} organizes the evaluated DG approaches according to their role in the learning pipeline, and Section~\ref{sec:related-work} reviews related work.
Section~\ref{sec:methods} describes the benchmark, experimental design, and evaluation protocol.
Section~\ref{sec:results-individual} and Section~\ref{sec:results-joint} present the individual and joint analyses of DG components, including their robustness across models and shifts and the available checkpoint-selection headroom.
Section~\ref{sec:discussion} discusses the broader implications, limitations, and directions for future work.
Finally, Section~\ref{sec:conclusion} summarizes the main findings.

\section{The Domain Generalization Problem}
\label{sec:dg-problem}

In ML, a domain ($\mathcal{S}$) characterizes a data-generating setting through the observation space and its underlying probability distribution~\cite{deeplearningbook}. 
Domain Generalization (DG) aims to train models that generalize to unseen domains, \ie, data collected under conditions different from those observed during training. 
Typically, models are trained in a supervised manner on multiple related source domains~\footnote{We use the term ``related'' to indicate that the source domains share the same task and label space.} and evaluated on a target domain unavailable during training. 
For example, a model trained on cartoon, sketch, and painting images of dogs (source domains) should perform well on real-world dog images (target domain). 
This setting is particularly relevant in practice, where distribution shifts can substantially degrade model performance~\cite{napoli2024benchmark,wang2022generalizing}.

Formally, let $\mathcal{X} \subseteq \mathbb{R}^{d}$ denote the input space and $\mathcal{Y} \subset \mathbb{Z}$ the label space. 
Let random variables $X \in \mathcal{X}$ and $Y \in \mathcal{Y}$ follow a joint distribution $P(X,Y)$. 
A domain $\mathcal{S}$ is defined as the triplet:
\begin{equation*}
    \mathcal{S} = \bigl\{P(X,Y),\,\mathcal{X},\,\mathcal{Y}\bigr\},
\end{equation*}
where $P(X,Y)$ denotes the probability distribution over inputs and labels.  
Following Gulrajani and Lopez-Paz~\cite{gulrajanisearch} and Wang \etal~\cite{wang2022generalizing}, we adopt the compact notation $P_{\mathcal{XY}}$ for this joint distribution.

A dataset of $\mathcal{S}$ is a finite collection of $n$ \iid samples drawn from $P_{\mathcal{XY}}$ so that:
\begin{equation*}
 S = \{(x_i, y_i)\}_{i=1}^{n}  \stackrel{\text{i.i.d.}}{\sim} P_{\mathcal{XY}},
\end{equation*}
where $x_i \in \mathcal{X}$ is an input vector and $y_i \in \mathcal{Y}$ the corresponding label.

In this definition, $\mathcal{X}$ and $\mathcal{Y}$ characterize the data and prediction task of a domain.
However, it is important to note that numerical transformations such as unit conversion, normalization, or standardization do not necessarily constitute meaningful domain shifts, since they preserve the underlying signal and acquisition conditions, that is, the data-generating process. 
In contrast, changes in users, devices, sensor positions, environments, or acquisition protocols alter the data-generating process and therefore represent meaningful shifts.\footnote{Our cross-dataset and cross-position scenarios reflect such acquisition changes rather than preprocessing transformations.} Changing the semantic label space $\mathcal{Y}$ likewise defines a different task, even when the sensor observations remain unchanged.

In DG, the training set $\mathcal{S}_{\text{tr}}$ often consists of multiple source domains, each sampled from a distinct joint distribution $P_{\mathcal{XY}}^{(i)}$, while sharing the same label space $\mathcal{Y}$\footnote{We assume a closed label set, meaning no new classes appear in the target domain.}. Formally:
\begin{equation*}
    \mathcal{S}_{\text{tr}}  \;=\; \Bigl\{S^{(i)} = \bigl\{(x_j^{(i)},\,y_j^{(i)})\bigr\}_{j=1}^{n_i}  \stackrel{\text{i.i.d.}}{\sim} P_{\mathcal{XY}}^{(i)} \;\Bigm|\; i=1,\ldots,M \Bigr\}
\end{equation*}
where the $M$ source domains differ as $P_{\mathcal{XY}}^{(i)} \neq P_{\mathcal{XY}}^{(j)}$ for $i \neq j$.
The goal is to learn a predictive function $h: \mathcal{X} \to \mathcal{Y}$ that performs well on an unseen target domain $S_{\text{test}}$, sampled from a different (``unseen'') joint distribution $P_{\mathcal{XY}}^{(u)}$:
\begin{equation*}
S_{\text{test}}
= \{(x_k^{(u)},y_k^{(u)})\}_{k=1}^{n_u}
\stackrel{\text{i.i.d.}}{\sim} P_{\mathcal{XY}}^{(u)},
\qquad
P_{\mathcal{XY}}^{(u)} \neq P_{\mathcal{XY}}^{(i)},\ \forall i.
\end{equation*}

Although $P_{\mathcal{XY}}^{(u)}$ is inaccessible during training, it is assumed to share some latent structure with the source domains, enabling knowledge transfer~\cite{wang2022generalizing,vuong2025domain}.
In fact, this assumption is necessary for DG to be feasible; otherwise, there would be no basis for transferring knowledge from source to target domains~\cite{wang2022generalizing}. 
This also justifies using multiple source domains during training, as they provide diversity that helps the model learn generalizable features~\cite{vuong2025domain}. 

\subsection{Related Concepts}

Several related paradigms address distribution shifts but differ in data availability and task scope. 
Domain Adaptation uses some amount of target-domain samples during training to align source and target distributions. Meanwhile, Multi-Task Learning jointly learns predictive functions for multiple related tasks known at training time. 
Zero-Shot Learning (ZSL) and Open Set Recognition (OSR), in contrast, address changes in the label space: ZSL predicts unseen classes, often using semantic information, whereas OSR handles $\mathcal{Y}_{\text{test}} \supset \mathcal{Y}_{\text{train}}$ by classifying known samples and rejecting unknown ones. 
DG differs from these paradigms by assuming a fixed shared label space $\mathcal{Y}$ and no access to target-domain data during training, making it particularly relevant when post-deployment adaptation is infeasible, such as medical imaging across hospitals~\cite{jahanifar2025domain}, industrial fault detection~\cite{zhao2024domain}, re-identification across unseen cameras~\cite{luo2020generalizing}, autonomous driving~\cite{sanchez2023domain}, and sensor-based HAR across unseen users, devices, or sensor placements~\cite{napoli2025domain}.


\section{Distribution Shifts in Human Activity Recognition}
\label{sec:har-generalization-challenges}

Smartphone-based HAR is highly susceptible to distribution shifts caused by differences in users, devices, sensor placements, and acquisition protocols, often leading to substantial degradation in unseen domains~\cite{napoli2024benchmark,lu2025harood,napoli2025domain,napoli2026components}. Previous studies report drops of up to approximately $40\%$ under unseen users or datasets~\cite{hong2024crosshar,napoli2024benchmark}, motivating DG methods for HAR~\cite{cai2025towards}. 
Such degradation is frequently hidden by conventional \iid protocols, in which training and test samples are drawn from the same underlying distribution~\cite{napoli2024benchmark,da2026benchmarking}. 

In this work, we focus on two scenarios: 
\textit{cross-dataset} (CD), where multiple acquisition factors may change simultaneously, and 
\textit{cross-position} (CP), which isolates sensor placement while keeping subjects, device, and acquisition protocol fixed. 
To illustrate CD shifts, we use the DAGHAR benchmark~\cite{napoli2024benchmark}, which comprises six smartphone datasets. 
The DAGHAR benchmark applies a common preprocessing pipeline, including resampling, gravity removal, and unit normalization, reducing trivial differences between datasets. 
For CP, we use the four RealWorld placements: thigh, shin, upper arm, and waist, also preprocessed with the same pipeline.
Table~\ref{tab:datasets-summary} summarizes both scenarios\footnote{Each dataset is treated as a distinct domain. In CP, each RealWorld sensor placement defines a separate domain.}.

\begin{table}[!hptb]
    \centering
    \caption{Information about the datasets considered in the Cross-Dataset (CD) and Cross-Position (CP) scenarios. }
    \label{tab:datasets-summary}
    \begin{tabular}{@{}lllcc@{}}
        \toprule
        \textbf{Dataset} & \textbf{Acronym} & \textbf{Position} &
        \textbf{CD} & \textbf{CP} \\
        \midrule
 RealWorld (Shin)~\cite{sztyler2016body}      & RW-S & Shin      &     & \ck \\
 RealWorld (U. Arm)~\cite{sztyler2016body}    & RW-U & Upper Arm &     & \ck \\
 RealWorld (Thigh)~\cite{sztyler2016body}     & RW-T & Thigh     & \ck & \ck \\
 RealWorld (Waist)~\cite{sztyler2016body}     & RW-W & Waist     & \ck & \ck \\
 KuHAR~\cite{nahid2021ku}                     & KH   & Waist     & \ck &     \\
 UCI-HAR~\cite{reyes2016transition}           & UCI  & Waist     & \ck &     \\
 MotionSense~\cite{malekzadeh2019mobile}      & MS   & Pocket    & \ck &     \\
 WISDM~\cite{weiss2019smartphone}             & WDM  & Pocket    & \ck &     \\
        \bottomrule
    \end{tabular}
\end{table}

\begin{figure*}[!ht]
    \centering
    \begin{subfigure}[t]{0.495\textwidth}
        \centering
        \includegraphics[width=\linewidth]{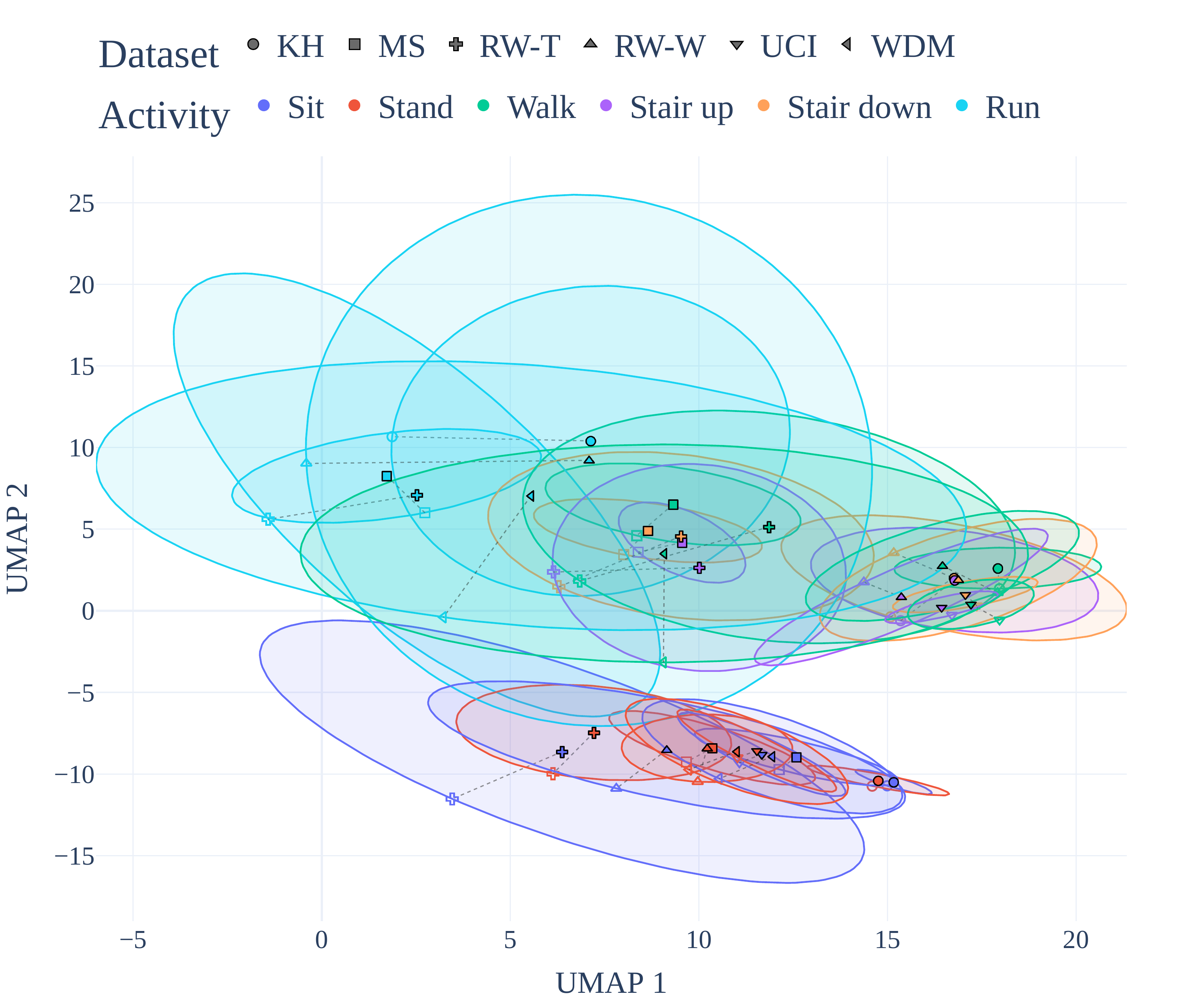}
        \caption{Cross-dataset shift.}
        \label{fig:crossdataset-shifts}
    \end{subfigure}
    \hfill
    \begin{subfigure}[t]{0.495\textwidth}
        \centering
        \includegraphics[width=\linewidth]{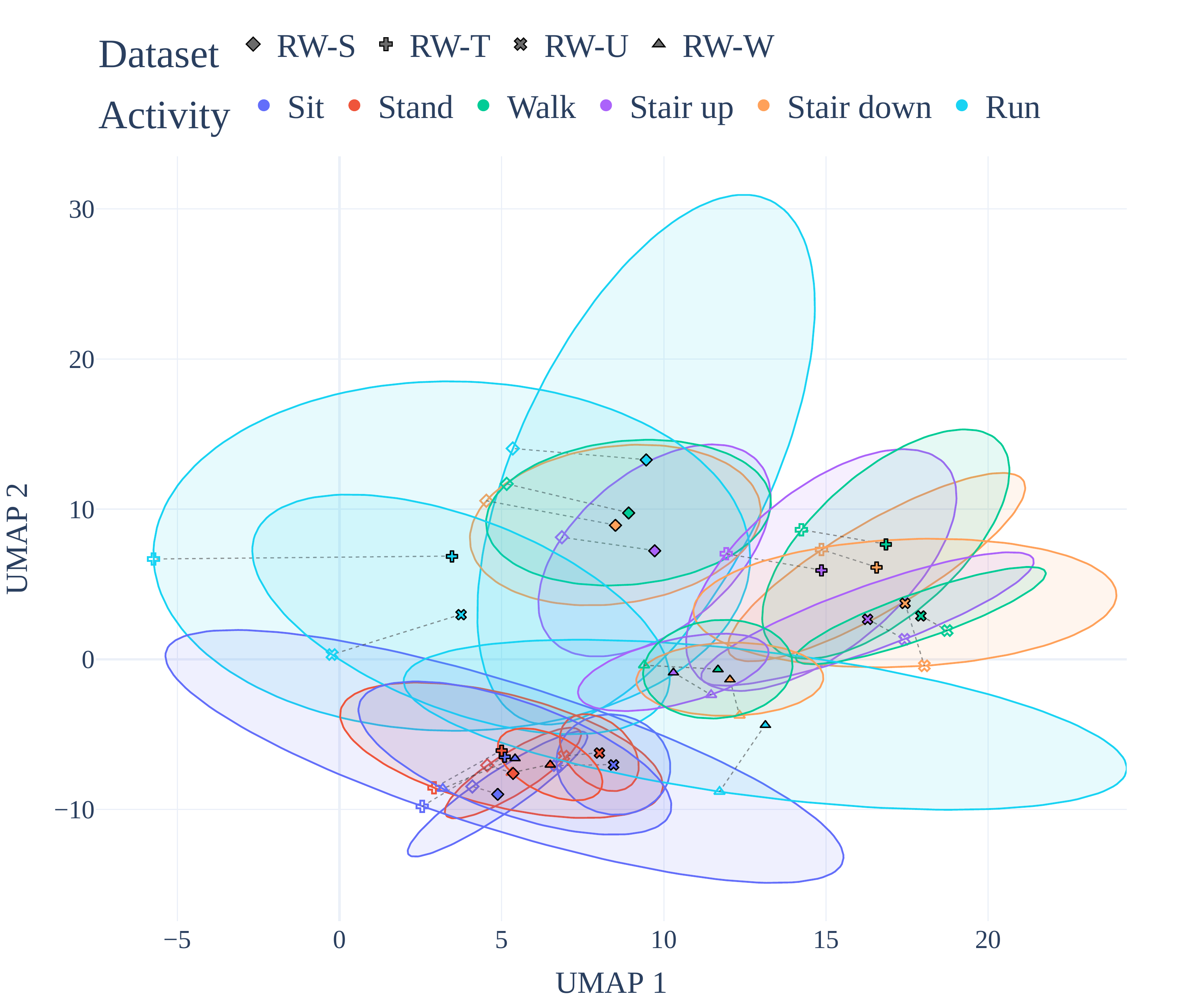}
        \caption{Cross-position shift.}
        \label{fig:crossposition-shifts}
    \end{subfigure}

    \caption{Qualitative visualization of CD and CP shifts. Inertial windows are transformed using FFT and projected using UMAP.  Colors denote activities, marker shapes denote domains, and ellipses represent two-standard-deviation covariance contours around domain--activity  centroids. Shaded markers relate the centroid to the corresponding ellipse. }
    \label{fig:frequency-shifts}
\end{figure*}

Figure~\ref{fig:frequency-shifts} qualitatively illustrates both scenarios using FFT representations projected with UMAP. 
Markers denote domain--activity centroids and ellipses their covariance contours.\footnote{Because UMAP is nonlinear, the visualization is qualitative and should not be interpreted as a direct measure of shift magnitude.}
Activity type remains the dominant source of organization, but same-activity distributions still vary across domains. 
In CD, locomotion activities, particularly walking and stair-related classes, exhibit clear dataset-dependent shifts that appear partly associated with sensor placement. 
CP makes this effect more explicit: even with subjects, device, and acquisition protocol fixed, changing body position substantially alters locomotion distributions, while static activities remain comparatively stable.

\subsection{Shift Magnitude and Potential Sources}
\label{sec:potential-sources-of-shift}

We quantify pairwise domain separability using four-fold cross-validated logistic regression on frequency-domain representations and report a normalized Proxy $\mathcal{A}$-Distance (PAD)~\cite{ben2006analysis}\footnote{For a domain-classification accuracy $a$, the conventional Proxy $\mathcal{A}$-Distance is $d_{\mathcal{A}}=4a-2$, ranging from $0$ to $2$ for accuracies at or above chance level. We report the normalized form $d_{\mathcal{A}}^{\mathrm{norm}}=\max(0,2a-1)$, bounded to $[0,1]$. A value of $0$ indicates no detectable domain separability, whereas $1$ indicates perfect separability. Below-chance accuracies arising from finite-sample variation are mapped to $0$.}. The domain classifier is trained within a cross-validation pipeline in which feature standardization is fitted only on the training folds, and the two domains are sampled with equal numbers of examples to avoid domain-class imbalance.

\begin{figure*}[!ht]
    \centering

    \begin{subfigure}[t]{0.45\textwidth}
        \centering
        \includegraphics[width=1.0\linewidth,trim=0 0 0 0,clip]
        {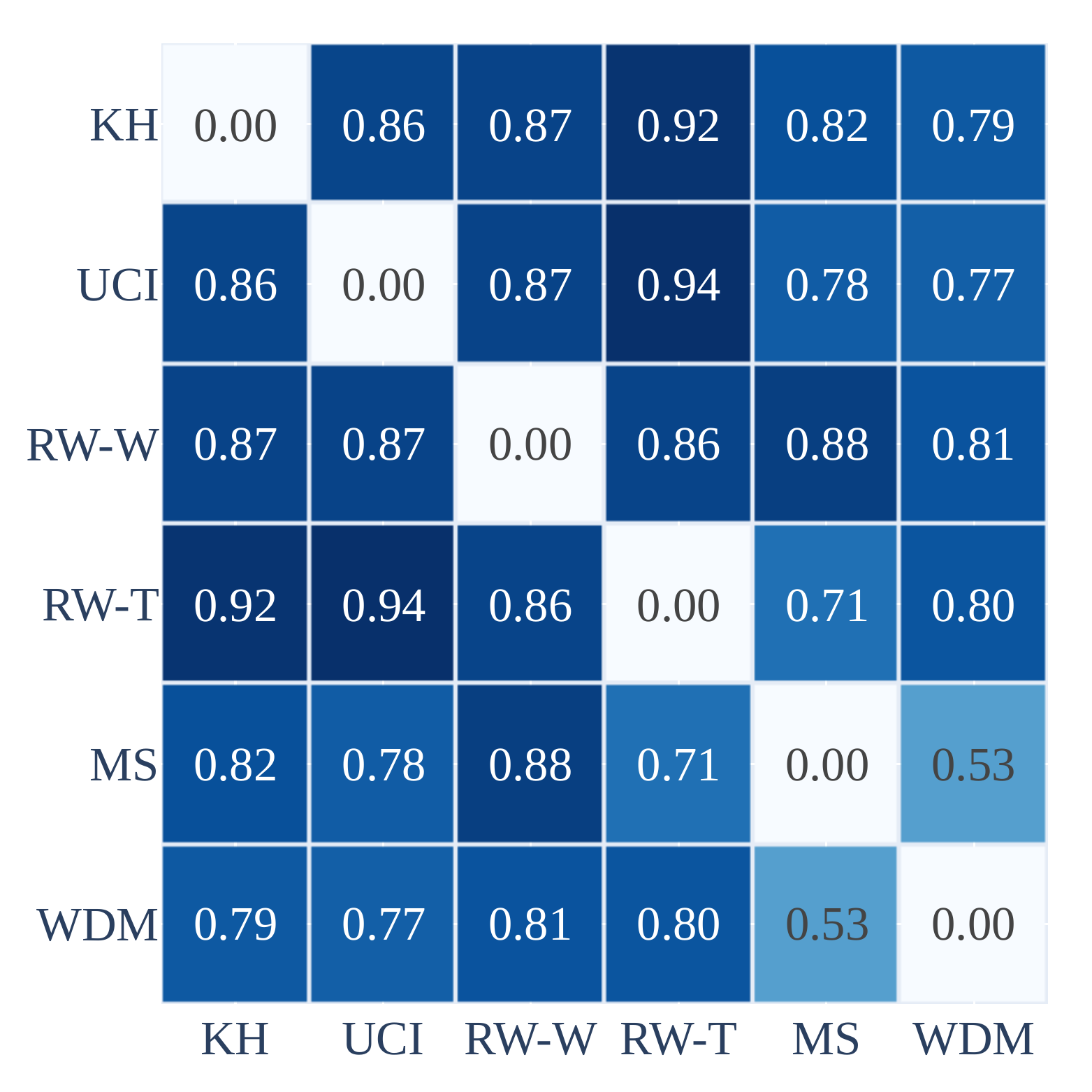}
        \caption{Cross-dataset shift.}
        \label{fig:crossdataset-pad}
    \end{subfigure}
    \hfill
    \begin{subfigure}[t]{0.45\textwidth}
        \centering
        \includegraphics[width=1.0\linewidth,trim=0 0 0 0,clip]
        {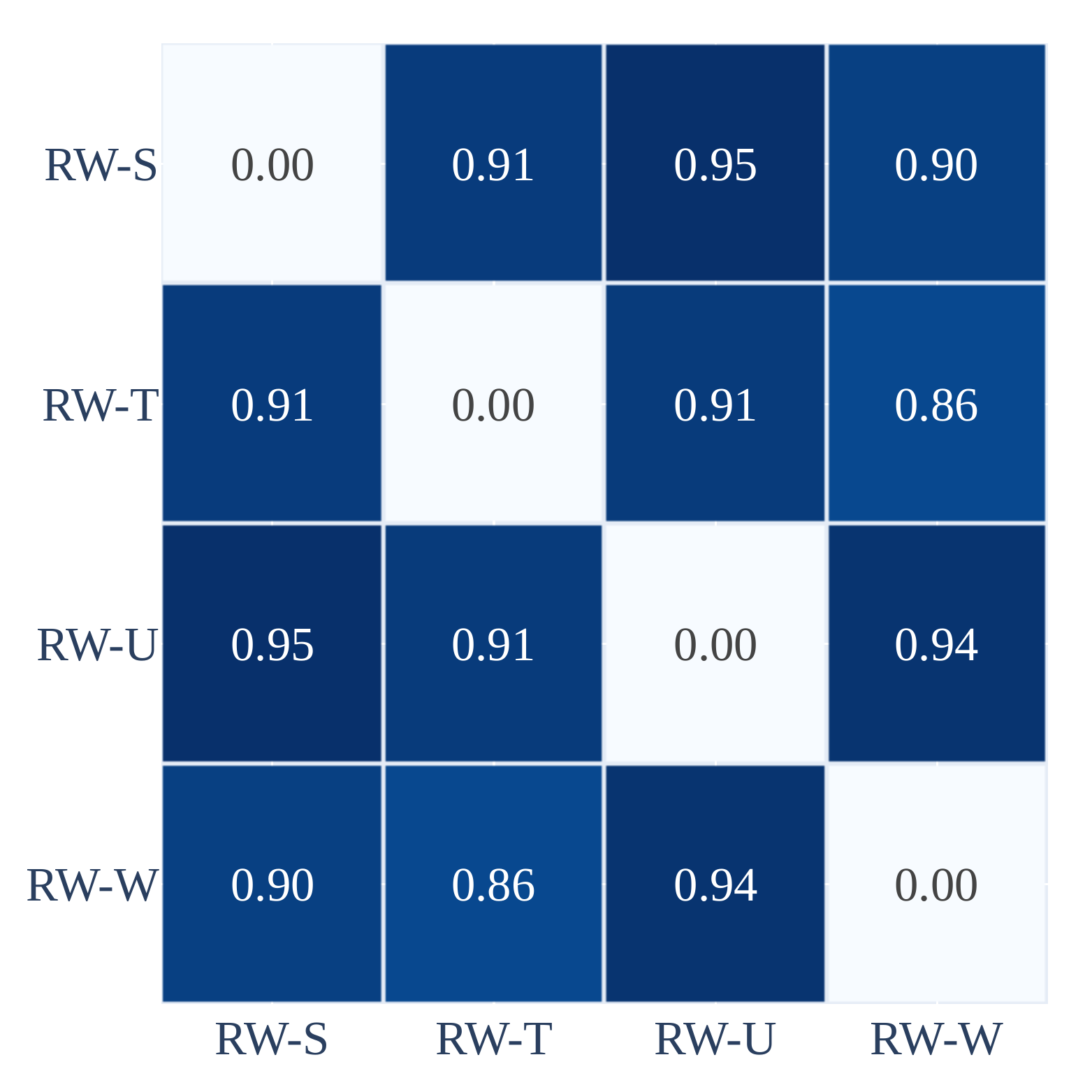}
        \caption{Cross-position shift.}
        \label{fig:crossposition-pad}
    \end{subfigure}
    \caption{Pairwise domain separability measured by four-fold cross-validated domain-classification accuracy, used as a proxy for $\mathcal{A}$-distance. Values closer to $1.0$ indicate greater domain separability, while $0.0$ indicates indistinguishable domains.}
    \label{fig:frequency-pad}
\end{figure*}

Figure~\ref{fig:frequency-pad} shows substantial separability in both scenarios: CD ranges from $0.53$ to $0.94$ (mean $0.814$), whereas CP ranges from $0.86$ to $0.95$ (mean $0.912$). Thus, changing sensor position alone can induce stronger distribution differences than the heterogeneous changes represented by CD.

To isolate potential sources of shift, we define a \textit{dataset family} as a common data-collection trial that shares subjects, devices, and acquisition protocol, regardless of position. Thus, all RealWorld placements form one family, while the remaining datasets form separate families. Across the $\binom{8}{2}=28$ pairs, we distinguish changes in position only, family only, or both.

Table~\ref{tab:factorial-shift} identifies sensor placement as the strongest source of separability: changing position within RealWorld yields a mean domain-classification accuracy of $0.912$, compared with $0.781$ when changing dataset family while approximately preserving position. Changing both yields $0.876$. 

\begin{table}[!ht]
    \centering
    \caption{Domain separability according to potential sources of shift across all $28$ dataset pairs. Values report the mean normalized Proxy $\mathcal{A}$-Distance using FFT features. Lower values indicate domain invariance.}
    \label{tab:factorial-shift}
    \begin{tabular}{lcc}
        \toprule
        \textbf{Source of variation} & \textbf{\begin{tabular}[c]{@{}c@{}}Mean\\accuracy ($\downarrow$)\end{tabular}} & \textbf{Pairs} \\ \midrule
        Different family, different position & $0.876$ & $18$ \\
        Different family, same position      & $0.781$ & $4$  \\
        Same family, different position      & $0.912$ & $6$  \\ 
        \bottomrule
    \end{tabular}
\end{table}

Hence, although CD combines more heterogeneous factors, CP systematically varies one of the strongest identifiable sources of shift, helping explain why it constitutes a particularly challenging generalization setting.


\section{Methods for Domain Generalization}
\label{sec:dg-methods}

A broad range of methods has been proposed to improve robustness under distribution shifts. 
Existing taxonomies commonly organize them by principles such as data manipulation, representation learning, and optimization~\cite{wang2022generalizing,lu2025harood}. 
However, recent evidence suggests that generalization depends on multiple interacting parts of the learning pipeline, including initialization, architecture, and optimization~\cite{teterwak2025erm++,teterwak2025large,napoli2025domain,napoli2026components}. Since HAR studies often evaluate these components in isolation~\cite{lu2025harood,hong2024crosshar}, it remains unclear whether gains arise from individual methods or from their interactions.

We therefore organize DG methods according to the component of the learning pipeline they modify:
\begin{enumerate}
    \item \textbf{Model initialization methods}: perform an additional pretraining stage, typically through self-supervised learning, to learn transferable representations before supervised DG training.
    
    \item \textbf{Architectural modification methods}: modify the model's internal computation by inserting or replacing operations within the feature extractor, like plug-in modules. These changes may be trainable or parameter-free, so their use depends on the backbone's internal structure rather than only on its training objective.

    \item \textbf{Objective-based methods}: modify supervised DG training via losses, regularization, gradients, sampling, or optimization, without altering the backbone's internal feature-extraction operations. The DG mechanism is introduced primarily through the training algorithm rather than persistent architectural changes.
\end{enumerate}

These categories are intentionally non-exclusive and can be combined within the same pipeline. For example, a pretrained model can be trained jointly with an architectural modification and an alternative DG objective. This functional taxonomy therefore enables controlled analysis of complementary, redundant, and model-dependent interactions.

Based on this taxonomy, Table~\ref{tab:dg-methods-summary} summarizes the representative methods considered in this work, spanning both modality-agnostic DG and HAR-specific approaches. ERM serves as the supervised baseline, while the remaining methods cover risk-based optimization, feature alignment, gradient regularization, feature augmentation and suppression, architectural adaptation, and self-supervised initialization. In particular, MixStyle, EFD-Mix, and DDG are treated as architectural modifications, whereas TF-C and LFR are used as representation initialization strategies.

\begin{table*}[!ht]
    \caption{
 Summary of the methods considered in this work. 
 Objective-based methods modify the training or optimization procedure; architectural modification methods introduce operations within the model; and model initialization methods provide pretrained representations before supervised DG training. 
 The ``Context'' column indicates the original setting in which each method was initially proposed or primarily evaluated. 
 Sup.: Supervised Learning; CV: computer vision; HAR: human activity recognition; TS: time-series. 
        \label{tab:dg-methods-summary}
 }
    \centering
    \small
    \setlength{\tabcolsep}{7pt}
    \renewcommand{\arraystretch}{1.18}

    \begin{tabular}{@{}p{0.13\textwidth}p{0.07\textwidth}p{0.74\textwidth}@{}}
        \toprule
        \textbf{Algorithm} &
        \textbf{Context} &
        \textbf{Core Idea} \\
        \midrule

        \multicolumn{3}{@{}l@{}}{\textit{Objective-Based Methods}} \\
        \midrule

 ERM~\cite{vapnik1998statistical} &
 Sup. &
 Minimizes the average classification loss over the aggregated source-domain samples. \\

 ERM++~\cite{teterwak2025erm++} &
 CV &
 Improves ERM through classifier warm-up, controlled fine-tuning, and parameter averaging. \\

 CORAL~\cite{sun2016deep} &
 CV &
 Aligns the means and covariances of feature distributions across source domains. \\

 GroupDRO~\cite{sagawa2019distributionally} &
 CV &
 Dynamically increases the weights of high-loss domains to improve worst-domain performance. \\

 VREx~\cite{krueger2021out} &
 CV &
 Penalizes differences between domain-specific risks to encourage consistent performance. \\

 Fish~\cite{shi2022gradient} &
 CV &
 Approximates inter-domain gradient alignment through sequential first-order meta-updates. \\

 FishR~\cite{rame2022fishr} &
 CV &
 Matches the variances of per-sample classifier gradients across source domains. \\

 RSC~\cite{huang2020self} &
 CV &
 Suppresses highly predictive feature dimensions and forces classification from alternative evidence. \\

 RDM~\cite{nguyen2024domain} &
 CV &
 Aligns distributions of per-sample risks rather than only domain-level average losses. \\

 LAG~\cite{lu2022local} &
 HAR &
 Aligns complementary local and global feature representations across source domains. \\

 FIXED~\cite{lu2024fixed} &
 CV/HAR &
 Combines domain-invariant feature Mixup, adversarial alignment, and large-margin classification. \\

 DIFEX~\cite{lu2022domain} &
 CV/HAR &
 Transfers frequency-phase knowledge while learning domain-invariant representations. \\

 SDMix~\cite{lu2022semantic} &
 HAR &
 Performs semantic-aware Mixup while encouraging large margins between activity classes. \\

        \midrule
        \multicolumn{3}{@{}l@{}}{\textit{Architectural Modification Methods}} \\
        \midrule

 MixStyle~\cite{zhou2021domain} &
 CV &
 Mixes channel-wise feature statistics to generate representations with synthetic domain styles. \\

 EFD-Mix~\cite{zhang2022exact} &
 CV &
 Interpolates complete feature distributions by matching and mixing sorted activations. \\

 DDG~\cite{sun2022dynamic} &
 CV &
 Generates input-dependent convolutional kernels by combining multiple learnable templates. \\

        \midrule
        \multicolumn{3}{@{}l@{}}{\textit{Model Initialization Methods}} \\
        \midrule

 TF-C~\cite{zhang2022self} &
 TS/HAR &
 Aligns time-domain and frequency-domain representations through self-supervised pretraining. \\

 LFR~\cite{sui2023self} &
 CV/HAR &
 Learns augmentation-free representations by predicting the outputs of multiple random projectors. \\
        \bottomrule
    \end{tabular}
\end{table*}

\section{Related Work}
\label{sec:related-work}

The effectiveness of DG methods is highly sensitive to the evaluation protocol. Differences in datasets, preprocessing, architectures, hyperparameter optimization, and checkpoint selection can substantially affect observed performance, making fair comparisons difficult.

Gulrajani and Lopez-Paz~\cite {gulrajanisearch} took a major step toward standardized DG evaluation with DomainBed. 
By controlling these factors within a unified benchmarking framework, DomainBed showed that many specialized DG methods do not consistently outperform ERM, emphasizing the importance of rigorous evaluation. 
Its protocol has since become the de facto standard adopted by numerous DG studies~\cite{lu2025harood,wang2022generalizing,teterwak2025erm++,napoli2026components}.

More recently, Teterwak \etal~\cite{teterwak2025erm++} revisited ERM under modern training practices and showed that improvements in representation initialization, optimization, and training procedures can substantially strengthen this baseline. 
Their ERM++ framework demonstrated that the effectiveness and ranking of DG methods depend not only on the DG objective itself, but also on the overall learning pipeline. 
This finding motivates evaluations that jointly consider training objectives, architectures, and representation initialization, rather than treating them as independent factors~\cite{napoli2026components}.

Compared with computer vision, comprehensive DG benchmarking remains less developed in HAR. 
Existing studies address subject, device, position, and dataset shifts~\cite{lu2022domain,lu2022semantic,hong2024crosshar}, but most evaluate newly proposed methods against a limited set of baselines in task-specific settings, making cross-study comparisons difficult.

Napoli \etal~\cite{napoli2024benchmark} took an important step toward standardized evaluation by introducing the DAGHAR preprocessing pipeline for smartphone-based HAR. 
Their study showed that standardizing datasets eliminates trivial differences in signal representation, allowing more reliable analysis of distribution shifts caused by users, devices, sensor placements, and acquisition protocols.
They also demonstrated that conventional \iid evaluation protocols can substantially overestimate generalization performance. However, the evaluation focused primarily on ERM.

Recently, Lu \etal~\cite{lu2025harood} introduced HAROOD, a DomainBed-inspired benchmark for sensor-based HAR that standardizes OOD evaluation across algorithms and backbone architectures and shows that performance can vary substantially with the chosen backbone and evaluation setting.

While HAROOD represents an important step toward standardized DG evaluation in HAR, it primarily compares standalone objective-based DG methods rather than systematically studying how different components of the learning pipeline interact. This is increasingly important given evidence that representation initialization, architectural design, and training objectives can jointly impact generalization performance~\cite{teterwak2025erm++,napoli2025domain,napoli2026components}. Moreover, HAROOD considers a relatively limited set of backbone architectures, does not evaluate self-supervised initialization or recent architectural DG mechanisms such as DDG, MixStyle, and EFD-Mix, and does not focus on smartphone-based HAR.
Thus, the individual and combined effects of these complementary DG components in modern smartphone-based HAR remain largely unexplored.

This work addresses this gap through, to the best of our knowledge, the first large-scale controlled evaluation in smartphone-based HAR that jointly studies representation initialization strategies, architectural DG modifications, and objective-based DG methods. By systematically crossing these components, we isolate their individual contributions and quantify their interactions and combined effects.

Building on standardized smartphone HAR datasets, DomainBed-inspired evaluation principles, and modern HAR architectures, we provide a unified analysis of how initialization, architecture, and training objectives influence the generalization performance.

\section{Materials and Methods}
\label{sec:methods}

All experiments follow a standardized protocol inspired by DomainBed~\cite{gulrajanisearch}. We use smartphone-based IMU data processed with the DAGHAR pipeline~\cite{napoli2024benchmark}, which harmonizes sampling rate, gravity removal, measurement units, and segmentation across datasets\footnote{This standardization minimizes artificial differences in signal representation, allowing the analysis to focus on distribution shifts caused by users, devices, sensor placements, and acquisition conditions.}.
Each sample is a $3$-second window sampled at $20$ Hz, yielding a $6 \times 60$ tensor containing tri-axial accelerometer and gyroscope signals\footnote{A 3-second window is adopted because durations between 2.5 and 3.5 seconds provide a favorable trade-off between recognition performance and latency for smartphone-based HAR~\cite{wang2018impact}.}.

\begin{figure*}[!hptb]
    \centering
    \includegraphics[width=0.90\textwidth]{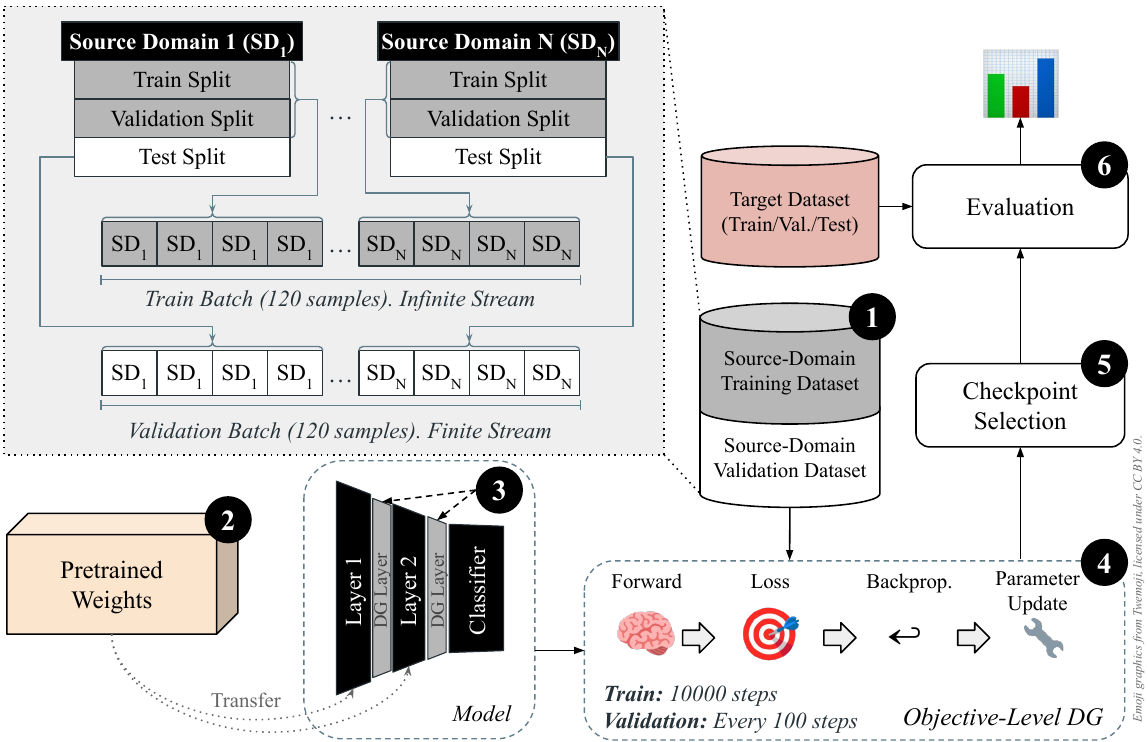}
    \caption{
 Overview of the DG protocol.
        \circnum{1} \textbf{Data partitioning:} one domain is held out as the unseen target, and the remaining domains are used for training and source validation.
        \circnum{2} \textbf{Model initialization:} the model is initialized randomly or through SSL pretraining.
        \circnum{3} \textbf{Architectural modification:} MixStyle, EFD-Mix, or DDG may be incorporated.
        \circnum{4} \textbf{DG objective:} training uses ERM or an alternative objective-based strategy.
        \circnum{5} \textbf{Training and selection:} models are trained for 10,000 steps, with checkpoints selected by mean source-validation accuracy.
        \circnum{6} \textbf{Target evaluation:} the selected checkpoint is evaluated on the held-out target domain.
        \label{fig:methodology}
 }
\end{figure*}

We consider the cross-dataset (CD) and cross-position (CP) scenarios described in Section~\ref{sec:har-generalization-challenges}. Figure~\ref{fig:methodology} summarizes the experimental protocol.

\subsection{Evaluation Protocol and Data Partitioning}
\label{sec:evaluation_protocol}

We adopt a strict leave-one-domain-out (LODO) protocol. 
CD comprises six folds: KuHAR, MotionSense, RealWorld-Thigh, RealWorld-Waist, UCI-HAR, and WISDM. In contrast, CP comprises four folds defined by the RealWorld shin, thigh, upper-arm, and waist positions (Table~\ref{tab:datasets-summary}). 
Unlike previous HAR DG benchmarks that combine smartphone and wearable datasets~\cite{lu2025harood,lu2022domain,lu2022semantic,hong2024crosshar,lu2024fixed}, our benchmark focuses exclusively on smartphone-based HAR.

Following the DAGHAR protocol~\cite{napoli2024benchmark}, we partition each domain by subject into training, validation, and test sets\footnote{This also enforces cross-subject evaluation within each domain, preventing participant overlap between training and evaluation.}. 
For source domains, we merge the training and validation partitions for supervised training (\circnum{1}) and use the test partition for source-domain validation and checkpoint selection. 
All target partitions are merged for final evaluation.\footnote{The target domain is excluded from pretraining, supervised optimization, hyperparameter tuning, and checkpoint selection.}

Training uses domain-balanced mini-batches of 120 samples (24 in CD and 40 in CP) with equal contribution from each source domain\footnote{Although larger batches can improve optimization stability, the scale of the experimental grid makes them substantially more expensive. A batch size of 120 therefore provides a practical compromise between training stability and computational cost.}.
Following DomainBed~\cite{gulrajanisearch}, domains are sampled as conceptually infinite streams, with exhausted samplers reshuffled and restarted.

\subsection{Models and DG Strategy Families}
\label{sec:models_methods}

We jointly vary representation initialization, architectural modification, and objective-based DG in a full-factorial design across model architectures, target domains, hyperparameters, and three random seeds, resulting in $410{,}400$ experiments.

We evaluate four representative HAR models: CNN-PFF~\cite{ha2016convolutional}, ResNet-SE-5~\cite{mekruksavanich2022deep}, TS2Vec~\cite{yue2022ts2vec}, and IMU-Transformer~\cite{shavit2021boosting}, spanning convolutional and transformer-based paradigms.

\paragraph{Initialization.} 
We consider Random, LFR (LODO), LFR (ES), TF-C (LODO), and TF-C (ES) (\circnum{2}). SSL initializes only the feature extractor, while the classification head remains random. LODO pretraining uses only source domains, whereas ES uses the external ExtraSensory dataset~\cite{vaizman2017recognizing}. For TF-C, only the time-domain encoder $G_t$ is transferred.

\paragraph{Architectural modifications.}
We compare the original model with MixStyle, EFD-Mix, and DDG (\circnum{3}). We always perform SSL pretraining before inserting the architectural modification.

\paragraph{Objectives.}
We evaluate thirteen objective-based strategies summarized in Table~\ref{tab:dg-methods-summary}. ERM is the reference objective, and Random initialization with the original architecture and ERM defines the benchmark baseline (``Plain ERM'')

\subsection{Training, Model Selection and Evaluation}
\label{sec:training_selection}

We train all supervised configurations for 10,000 optimization steps using Adam (\circnum{4}). 
A step-based budget ensures the same number of updates across domains and methods despite differences in dataset size. Appendix~\ref{sec:appendix-training_steps-ablation} shows that this budget is sufficient for convergence.

We search over learning rates $10^{-3}$ and $10^{-4}$ together with the method-specific grids reported in Appendix~\ref{sec:appendix-hparam-grids}. We train each configuration with three random seeds and evaluate it every 100 steps.

Checkpoint and hyperparameter selection rely exclusively on source-domain validation (\circnum{5}). For each run, we select the checkpoint with the highest mean source-validation accuracy, and we select hyperparameters based on mean validation accuracy across seeds. We then report target performance as the mean and standard deviation over the corresponding three runs.

For each candidate, we independently optimize all components not fixed by a particular analysis. 
Unlike the original DomainBed protocol, we report the single source-validation-selected configuration rather than averaging across multiple hyperparameter settings, better reflecting practical deployment.

We evaluate the selected models on the held-out target domain (\circnum{6}). Accuracy is the primary metric, as target domains have balanced class distributions, and we use paired Wilcoxon signed-rank tests with $p<0.05$ for statistical comparisons. We report differences in percentage points (pp).

\subsection{Research Questions}
\label{sec:research-questions}

We address four research questions:

\begin{itemize}
    \item RQ1: How much do representation initialization, architectural modifications, and objective-based DG methods individually contribute to unseen-domain generalization?

    \item RQ2: How do these components interact when combined, and when are their effects complementary, redundant, or conflicting?

    \item RQ3: How consistent are DG improvements across models and CD and CP shifts?

    \item RQ4: How reliably can source-domain validation select checkpoints that generalize to unseen domains, and how much performance remains under oracle selection?
\end{itemize}

\section{Individual Contributions of DG Components}
\label{sec:results-individual}

We first isolate each DG dimension. 
We evaluate objective-based methods with random initialization and the original model, initialization strategies under ERM with the original model, and architectural modifications under ERM with random initialization. 
We report results separately for cross-dataset (CD) and cross-position (CP).

\subsection{Objective-Based DG Methods}
\label{sec:results-objective-based-dg-methods-individual}

\begin{figure*}[!ht]
    \centering
    \includegraphics[width=\textwidth]{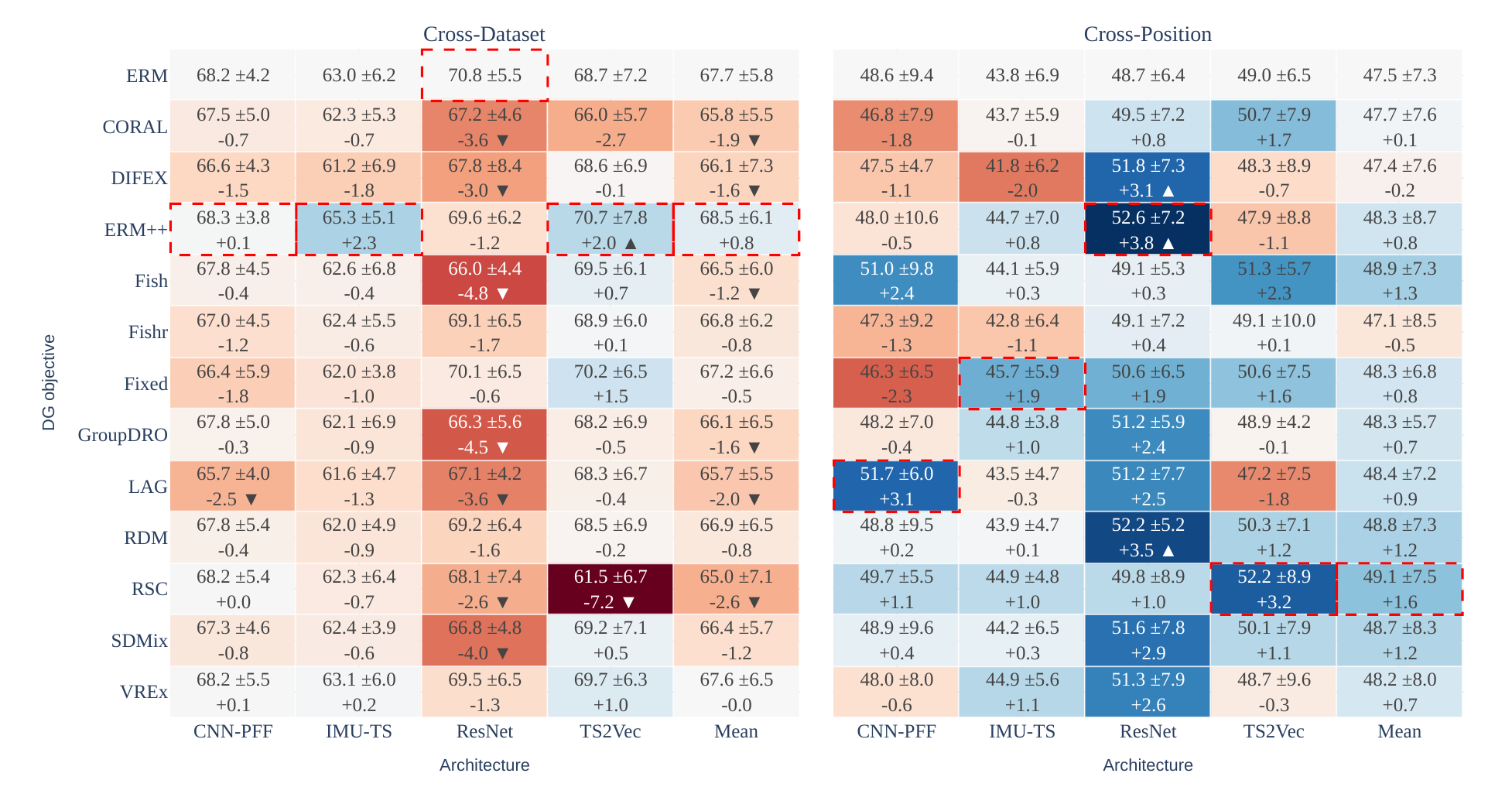}
    \caption{Performance of objective-based DG methods with random initialization and no architectural modification. Cells report mean target accuracy ($\pm$ standard deviation) and the difference relative to ERM. Color denotes the magnitude of the difference, while triangles indicate statistical significance (upward for improvement, downward for degradation), and dashed boxes the best value in each column. 
 }
 \label{fig:heatmap_best_objective}
\end{figure*}

Figure~\ref{fig:heatmap_best_objective} reports the source-validation-selected results. 
In CD, no alternative objective significantly outperforms ERM, while several significantly degrade aggregate performance. 
ERM++ is the most competitive alternative, with the largest positive mean difference ($+0.8$~pp), though it is not statistically significant. 
In CP, no objective differs significantly from ERM at the aggregate level.

The effect remains strongly model-dependent: CD degradations are concentrated mainly on ResNet-SE-5, whereas ERM++ improves IMU-Transformer and TS2Vec by $+2.3$ and $+2.0$~pp, respectively. 
Across the hyperparameter space explored (Table~\ref{tab:objective_variability}), ERM++ is the most robust alternative, with $52.9\%$ of runs outperforming plain ERM.
Replacing ERM alone therefore provides limited and highly conditional gains, consistent with previous DG evaluations~\cite{lu2025harood,teterwak2025erm++,gulrajanisearch}.

\begin{table}[!ht]
    \centering
    \caption{Robustness of objective-based DG methods across the complete hyperparameter search space. Positive rate ($Pos.$) is the percentage of runs outperforming plain ERM, and $\Delta$ is the mean target-accuracy difference in pp. Pooled values combine CD and CP.}
    \label{tab:objective_variability}
    \setlength{\tabcolsep}{3pt}
    \begin{tabular}{lrrrrrr}
        \toprule
        & \multicolumn{2}{c}{\textbf{CD}} &
        \multicolumn{2}{c}{\textbf{CP}} &
        \multicolumn{2}{c}{\textbf{Pooled}} \\
        \cmidrule(lr){2-3}\cmidrule(lr){4-5}\cmidrule(lr){6-7}
        \textbf{Objective} &
        \textbf{Pos. (\%)} & $\Delta$\textbf{ (pp)} &
        \textbf{Pos. (\%)} & $\Delta$\textbf{ (pp)} &
        \textbf{Pos. (\%)} & $\Delta$\textbf{ (pp)} \\
        \midrule
 CORAL    & 32.9 & -2.0 & 52.1 & +0.0 & 40.6 & -1.2 \\
 DIFEX    & 23.2 & -2.9 & 51.9 & +0.2 & 34.7 & -1.7 \\
 ERM++    & 47.9 & -0.8 & 60.4 & +1.1 & 52.9 & 0.0 \\
 Fish     & 35.1 & -1.8 & 62.2 & +1.4 & 45.9 & -0.5 \\
 FishR    & 30.6 & -2.3 & 54.0 & +0.3 & 39.9 & -1.2 \\
 Fixed    & 40.1 & -2.2 & 50.2 & -0.3 & 44.2 & -1.4 \\
 GroupDRO & 31.4 & -1.6 & 52.7 & +0.4 & 39.9 & -0.8 \\
 LAG      & 32.2 & -2.1 & 48.8 & -0.3 & 38.8 & -1.4 \\
 RDM      & 23.9 & -3.5 & 56.0 & +0.7 & 36.8 & -1.8 \\
 RSC      & 31.8 & -3.8 & 49.8 & -0.3 & 39.0 & -2.4 \\
 SDMix    & 42.2 & -1.6 & 54.0 & +0.3 & 46.9 & -0.8 \\
 VREx     & 21.2 & -7.0 & 52.1 & -1.0 & 33.6 & -4.6 \\
        \bottomrule
    \end{tabular}%
\end{table}

\subsection{Representation Initialization}
\label{sec:results-initialization-individual}

\begin{figure*}[!ht]
    \centering
    \includegraphics[width=\textwidth]{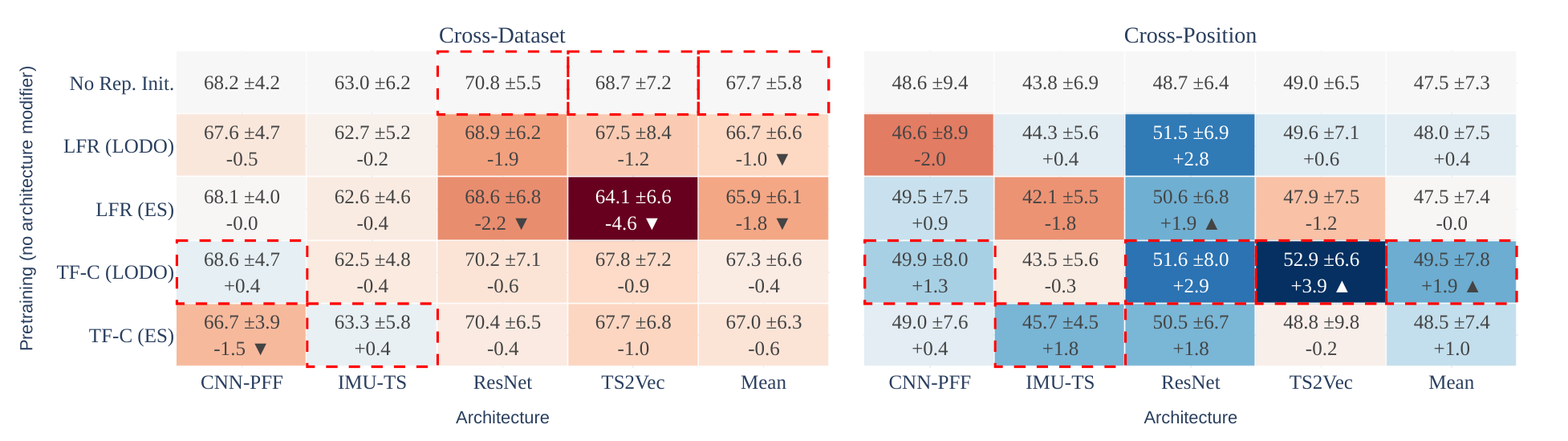}
    \caption{Effect of representation initialization under ERM and no architectural modification. Cells report mean target accuracy ($\pm$ standard deviation) and the difference relative to random initialization. Color denotes the magnitude of the difference, while triangles indicate statistical significance (upward for improvement, downward for degradation), and dashed boxes indicate the best value in each column.}
    \label{fig:heatmap_ssl_individual}
\end{figure*}

Figure~\ref{fig:heatmap_ssl_individual} evaluates LFR and TF-C pretraining under ERM. SSL provides limited aggregate gains in CD, whereas TF-C (LODO) significantly improves CP by $+1.9$~pp, including gains of $+2.9$~pp for ResNet-SE-5 and $+3.9$~pp for TS2Vec.

Pretraining effectiveness is also model- and scenario-dependent, with external ExtraSensory pretraining particularly unfavorable for TS2Vec in CD. 
In contrast, TF-C (LODO) is the only initialization that consistently improves performance across models and scenarios, achieving positive rates of up to $55\%$ under pooled evaluation (Table~\ref{tab:ssl_variability}) across the complete hyperparameter search space.
This suggests that alignment between pretraining and downstream source distributions is important, consistent with observations in prior works~\cite{teterwak2025large,da2026systematic}.

\begin{table}[!hptb]
\centering
\caption{Robustness of representation initialization under ERM. Positive rate ($Pos.$) denotes the percentage of runs outperforming random initialization, and $Delta$ is the mean target-accuracy difference in pp.}
\label{tab:ssl_variability}
\setlength{\tabcolsep}{3pt}
\begin{tabular}{lrrrrrr}
\toprule
& \multicolumn{2}{c}{\textbf{CD}} &
  \multicolumn{2}{c}{\textbf{CP}} &
  \multicolumn{2}{c}{\textbf{Pooled}} \\
\cmidrule(lr){2-3}\cmidrule(lr){4-5}\cmidrule(lr){6-7}
\textbf{Initialization} &
\textbf{Pos. (\%)} & $\Delta$\textbf{ (pp)} &
\textbf{Pos. (\%)} & $\Delta$\textbf{ (pp)} &
\textbf{Pos. (\%)} & $\Delta$\textbf{ (pp)} \\
\midrule
LFR (ES)       & 49.3 & -0.8 & 56.2 & +0.1 & 52.1 & -0.4 \\
LFR (LODO)     & 45.8 & -0.4 & 56.2 & +0.1 & 50.0 & -0.2 \\
TF-C (ES)      & 42.4 & -1.5 & 57.3 & +0.4 & 48.3 & -0.7 \\
TF-C (LODO)    & 48.6 & -0.4 & 64.6 & +1.5 & 55.0 & +0.4 \\
\bottomrule
\end{tabular}%
\end{table}

\subsection{Architectural Modifications}
\label{sec:results-architectural-modifications-individual}

\begin{figure*}[!hptb]
    \centering
    \includegraphics[width=\textwidth]{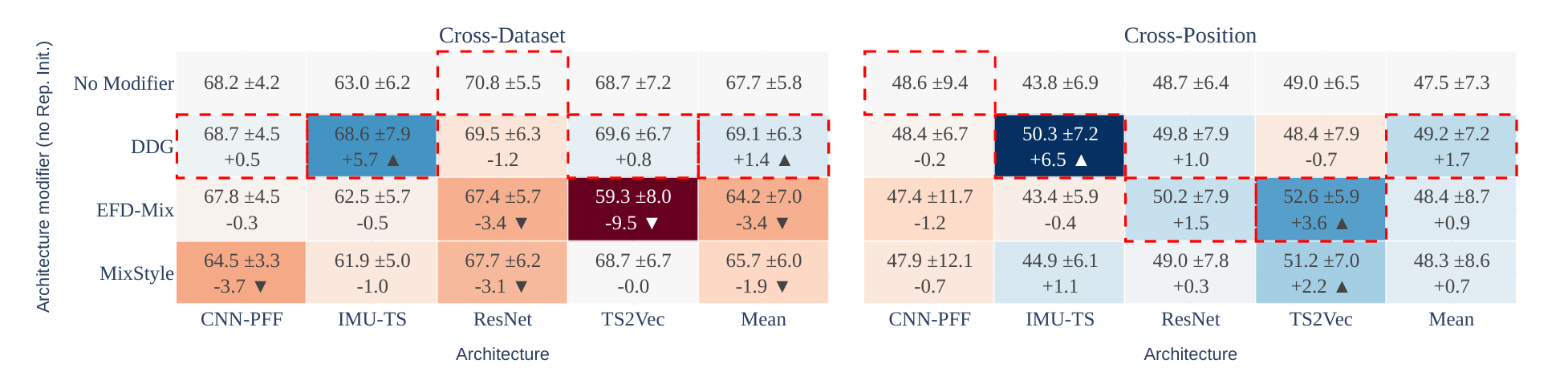}
    \caption{Effect of architectural modifications under ERM and random initialization. Cells report mean target accuracy ($\pm$ standard deviation) and the difference relative to the unmodified model. Color denotes the magnitude of the difference, while triangles indicate statistical significance (upward for improvement, downward for degradation), and dashed boxes indicate the best value in each column.}
    \label{fig:heatmap_modifier_individual}
\end{figure*}

Figure~\ref{fig:heatmap_modifier_individual} identifies DDG as the strongest standalone architectural modification, significantly improving aggregate CD accuracy by $+1.4$~pp.
In fact, the strongest interaction is observed between DDG and IMU-Transformer, yielding gains of $+5.7$~pp in CD and $+6.5$~pp in CP, both statistically significant.
In contrast, EFD-Mix and MixStyle can substantially degrade the performance of specific models, reducing the aggregate CD accuracy by $3.4$ and $1.9$~pp, respectively.

DDG also stands out in terms of robustness (Table~\ref{tab:modifier_variability}), with $57.5\%$ of runs outperforming plain ERM and positive mean gains across all scenarios. 
In fact, DDG is the only component that consistently improves performance in both scenarios while also achieving the largest mean gains, suggesting that it is the most reliable architectural modification for smartphone-based HAR.

\begin{table}[!hptb]
\centering
\caption{Robustness of architectural modifications under ERM with random initialization. Positive rate ($Pos.$) denotes the percentage of runs outperforming the unmodified model, and $\Delta$ is the mean target-accuracy difference in pp.}
\label{tab:modifier_variability}
\setlength{\tabcolsep}{3pt}
\begin{tabular}{lrrrrrr}
\toprule
& \multicolumn{2}{c}{\textbf{CD}} &
  \multicolumn{2}{c}{\textbf{CP}} &
  \multicolumn{2}{c}{\textbf{Pooled}} \\
\cmidrule(lr){2-3}\cmidrule(lr){4-5}\cmidrule(lr){6-7}
\textbf{Modifier} &
\textbf{Pos. (\%)} & $\Delta$\textbf{ (pp)} &
\textbf{Pos. (\%)} & $\Delta$\textbf{ (pp)} &
\textbf{Pos. (\%)} & $\Delta$\textbf{ (pp)} \\
\midrule
DDG      & 56.9 & +1.2 & 58.3 & +1.1 & 57.5 & +1.2 \\
EFD-Mix  & 27.1 & -3.9 & 60.4 & +0.9 & 40.4 & -2.0 \\
MixStyle & 37.2 & -1.3 & 57.3 & +0.6 & 45.2 & -0.6 \\
\bottomrule
\end{tabular}%
\end{table}

\subsection{Summary of Individual DG Components}
\label{sec:results-individual-components-summary}

Overall, no individual DG component is universally superior. Objective-based methods provide the weakest standalone gains, with ERM++ as the most competitive alternative to ERM. 
TF-C (LODO) is the most reliable initialization, while DDG provides the clearest architectural gains. 
Across all three families, effectiveness is strongly model- and shift-dependent, with CP generally more receptive to DG interventions than CD.
These results motivate evaluating DG components jointly rather than inferring their usefulness from isolated performance.

\section{Joint Contributions of DG Components}
\label{sec:results-joint}

We now analyze how the three DG dimensions interact when combined within the same pipeline.

\subsection{Overall Performance Landscape}
\label{sec:results-overall-landscape}

Table~\ref{tab:rq3a_top10_combined} summarizes the ten best source-validation-selected configurations for CD and CP.\footnote{We use the ten best configurations to capture the strongest DG solutions without relying on a single winner or the entire search space. Appendix~\ref{sec:appendix-number-of-configurations} analyzes this choice.} Configurations are ranked by mean target accuracy, with improvements measured relative to plain ERM.

Joint configurations provide substantially larger gains than individual DG components. In CD, all ten leading configurations include DDG, with ERM++ + DDG achieving the best result ($70.5\%$, $+2.9$~pp over ERM). 
In CP, the best configuration combines ERM++, LFR (LODO), and DDG ($+4.9$~pp), while nine of the top ten configurations include DDG and nine use SSL initialization. These results show that components with limited standalone gains can become effective when combined, motivating the complementarity analysis that follows.

\begin{table*}[!hptb]
\caption{Ten best joint configurations for CD and CP. Accuracy is averaged across target domains, models, and random seeds; improvements over plain ERM are reported in percentage points (pp) in parentheses. {\scriptsize $\blacktriangle$} denotes a statistically significant improvement over plain ERM.}
\label{tab:rq3a_top10_combined}
\centering
\small
\begin{tabular}{llll@{\hspace{1.2em}}llll}
\toprule
\multicolumn{4}{c}{\textbf{\textit{Cross-Dataset}}} &
\multicolumn{4}{c}{\textbf{\textit{Cross-Position}}} \\
\cmidrule(lr){1-4}\cmidrule(lr){5-8}
\textbf{DG Obj.} &
\textbf{Initialization} &
\textbf{Modifier} &
\textbf{Accuracy} &
\textbf{DG Obj.} &
\textbf{Initialization} &
\textbf{Modifier} &
\textbf{Accuracy} \\
\cmidrule(lr){1-4}\cmidrule(lr){5-8}

ERM++ & Random       & DDG & 70.5 (+2.9{\scriptsize $\blacktriangle$}) &
ERM++ & LFR (LODO)  & DDG & 52.5 (+4.9{\scriptsize $\blacktriangle$}) \\

ERM++ & LFR (LODO)  & DDG & 70.2 (+2.5{\scriptsize $\blacktriangle$}) &
VREx  & LFR (LODO)  & DDG & 51.8 (+4.2{\scriptsize $\blacktriangle$}) \\

ERM++ & TF-C (LODO) & DDG & 70.1 (+2.4{\scriptsize $\blacktriangle$}) &
SDMix & TF-C (LODO) & DDG & 51.8 (+4.2{\scriptsize $\blacktriangle$}) \\

ERM++ & LFR (ES)    & DDG & 69.7 (+2.1{\scriptsize $\blacktriangle$}) &
ERM++ & TF-C (LODO) & DDG & 51.8 (+4.2{\scriptsize $\blacktriangle$}) \\

Fixed & Random      & DDG & 69.4 (+1.7{\scriptsize $\blacktriangle$}) &
ERM++ & TF-C (ES)   & DDG & 51.7 (+4.2{\scriptsize $\blacktriangle$}) \\

ERM++ & TF-C (ES)   & DDG & 69.2 (+1.6) &
ERM++ & LFR (ES)    & DDG & 51.7 (+4.2{\scriptsize $\blacktriangle$}) \\

FishR & Random      & DDG & 69.1 (+1.5) & 
SDMix & LFR (ES)    & DDG & 51.6 (+4.1{\scriptsize $\blacktriangle$}) \\

Fixed & TF-C (LODO)  & DDG & 69.1 (+1.4{\scriptsize $\blacktriangle$}) &
ERM++ & Random       & DDG & 51.5 (+4.0{\scriptsize $\blacktriangle$}) \\

ERM   & Random       & DDG & 69.1 (+1.4{\scriptsize $\blacktriangle$}) &
Fish  & TF-C (LODO)  & DDG & 51.4 (+3.8{\scriptsize $\blacktriangle$}) \\

RDM   & Random       & DDG & 69.0 (+1.3) &
RSC   & TF-C (ES)    & EFD-Mix & 51.4 (+3.8{\scriptsize $\blacktriangle$}) \\

\bottomrule
\end{tabular}
\end{table*}

\subsection{Complementarity Across DG Dimensions}
\label{sec:results-complementarity}

\begin{figure*}[!hptb]
    \centering
    \includegraphics[width=1.0\linewidth]{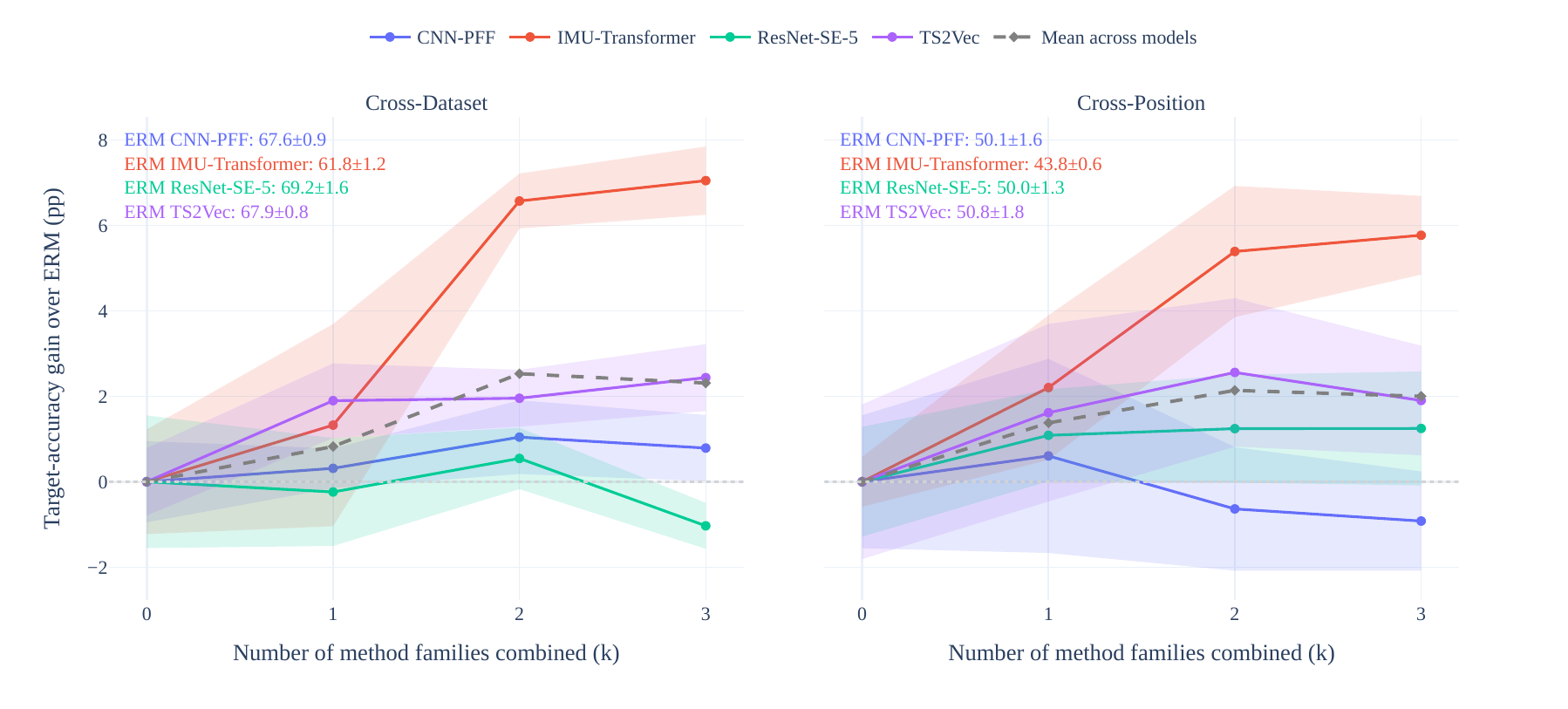}
    \caption{Complementarity across DG dimensions. For each model, scenario, and number $k$ of active dimensions, the figure reports the mean improvement over plain ERM for the ten best source-validation-selected configurations. Solid lines represent individual models, the dashed line the mean across models, and the shaded region the corresponding standard deviation.}
    \label{fig:complementarity_ladder}
\end{figure*}

We next analyze performance as the number of active DG dimensions increases. 
We define 
$k=0$ as plain ERM; 
$k=1$ activates one of initialization, architectural modification, or objective-based DG; 
$k=2$ combines two; and 
$k=3$ combines all three. 
For each model, scenario, and $k$, Figure~\ref{fig:complementarity_ladder} reports the mean improvement of the ten best source-validation-selected configurations over $k=0$.

On average, CD improves from $+0.82$~pp at $k=1$ to $+2.52$~pp at $k=2$, before slightly decreasing to $+2.31$~pp at $k=3$. 
CP follows a similar pattern ($+1.37$, $+2.13$, and $+2.00$~pp). Thus, combining DG dimensions is generally beneficial, but activating all three does not necessarily outperform the best two-dimensional combinations.

The trajectories are strongly model-dependent. 
IMU-Transformer exhibits the clearest complementarity, reaching $+7.05$~pp in CD and $+5.77$~pp in CP at $k=3$. 
TS2Vec also benefits from additional dimensions, whereas CNN-PFF and ResNet-SE-5 show weaker or negative effects in some settings. 
This heterogeneity indicates that the value of additional DG components depends on their compatibility with the underlying model.

\begin{figure*}[!hptb]
    \centering
    \includegraphics[width=1.0\linewidth]{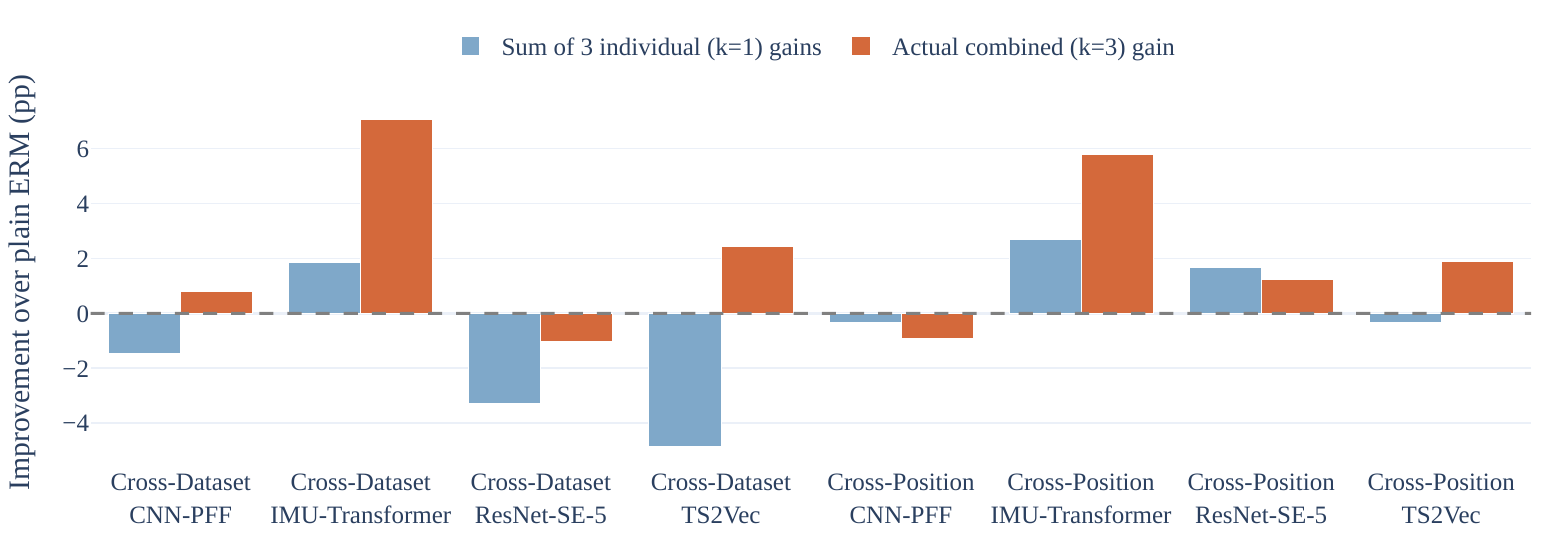}
    \caption{Empirical interaction among DG dimensions. We compare observed improvements of the top-$10$ configurations using all three dimensions with the sum of their isolated effects. Positive gaps indicate super-additive interaction, and negative gaps indicate redundancy or interference. We do not interpret cases where both quantities are negative as synergy.}
    \label{fig:superadditivity}
\end{figure*}

To distinguish interaction from simple accumulation, Figure~\ref{fig:superadditivity} compares the observed gain of the top-$10$ $k=3$ configurations with the sum of their isolated component effects. 
Positive gaps indicate empirical superadditivity, whereas negative gaps indicate redundancy or interference.

Five of the eight model--scenario pairs exhibit super-additive behavior. 
The strongest occurs for IMU-Transformer in CD, where isolated effects sum to only $+1.86$~pp but the joint configurations reach $+7.05$~pp. 
TS2Vec in CD is also notable. Despite isolated effects summing to $-4.82$~pp, their combination improves ERM by $+2.44$~pp. 
ResNet-SE-5 is the main exception.

Overall, the results show that DG components frequently provide complementary gains, but these interactions are neither universal nor monotonic with the number of active dimensions. 
Performance therefore depends more on selecting compatible combinations than simply adding more DG mechanisms.


\subsection{Component Attribution of Winning Configurations}
\label{sec:results-winning-component-attribution}

We next attribute the gains of the single best source-validation-selected pipeline for each model--scenario pair to its objective, initialization strategy, and architectural modification. 
Unlike the preceding analyses, which summarize the ten best configurations, this analysis focuses on the single best pipeline for each model--scenario pair. 
We decompose each selected pipeline into its individual components, pairwise combinations, and complete configuration, as summarized in Table~\ref{tab:winning_component_attribution}.

\begin{table*}[!hptb]
\centering
\caption{Component attribution for the winning configuration of each model and scenario. For each selected recipe, the objective-based method, initialization strategy, and architectural modification are evaluated individually, in pairs, and jointly in the complete configuration (``Full''). Values report mean accuracy $\pm$ standard deviation across target domains and random seeds, with the improvement over plain ERM shown in parentheses in percentage points (pp). Bold indicates the best result for each model and scenario.}
\label{tab:winning_component_attribution}
\resizebox{\textwidth}{!}{%
\setlength{\tabcolsep}{2.5pt}
\begin{tabular}{clllllccccccc}
\toprule
\multicolumn{1}{l}{} &
  \textbf{Model} &
  \textbf{DG Obj.} &
  \textbf{Modifier} &
  \textbf{Initialization} &
  \textbf{Baseline} &
  \textbf{Obj. Only} &
  \textbf{Modif. only} &
  \textbf{SSL only} &
  \textbf{Obj. + Modif.} &
  \textbf{Obj. + SSL} &
  \textbf{Modif. + SSL} &
  \textbf{Full} \\ \midrule
\multirow{4}{*}{\rotatebox{90}{\textbf{Cross-Domain}}} &
 CNN-PFF &
 ERM++ &
 DDG &
 LFR (ES) &
 68.2 $\pm$ 4.2 &
  \shortstack[c]{68.3 $\pm$ 3.6 \\ (+0.1)} &
  \shortstack[c]{68.7 $\pm$ 4.4 \\ (+0.5)} &
  \shortstack[c]{68.1 $\pm$ 4.0 \\ (-0.0)} &
  \shortstack[c]{69.2 $\pm$ 3.6 \\ (+1.1)} &
  \shortstack[c]{68.8 $\pm$ 4.2 \\ (+0.7)} &
  \shortstack[c]{68.2 $\pm$ 4.6 \\ (+0.0)} &
  \textbf{\shortstack[c]{70.2 $\pm$ 4.8 \\ (+2.1)}} \\
 &
 IMU-TS &
 ERM++ &
 DDG &
 Random &
 63.0 $\pm$ 6.0 &
  \shortstack[c]{65.3 $\pm$ 5.0 \\ (+2.3)} &
  \shortstack[c]{68.6 $\pm$ 8.1 \\ (+5.7)} &
  \shortstack[c]{63.0 $\pm$ 6.0 \\ (+0.0)} &
  \shortstack[c]{72.5 $\pm$ 8.9 \\ (+9.5)} &
  \shortstack[c]{65.3 $\pm$ 5.0 \\ (+2.3)} &
  \shortstack[c]{68.6 $\pm$ 8.1 \\ (+5.7)} &
  \textbf{\shortstack[c]{72.5 $\pm$ 8.9 \\ (+9.5)}} \\
 &
 ResNet &
 Fixed &
 DDG &
 TF-C (LODO) &
 70.8 $\pm$ 5.2 &
  \shortstack[c]{70.1 $\pm$ 6.3 \\ (-0.6)} &
  \shortstack[c]{69.5 $\pm$ 6.4 \\ (-1.2)} &
  \shortstack[c]{70.2 $\pm$ 7.3 \\ (-0.6)} &
  \shortstack[c]{71.8 $\pm$ 7.2 \\ (+1.0)} &
  \shortstack[c]{71.7 $\pm$ 5.3 \\ (+0.9)} &
  \shortstack[c]{70.9 $\pm$ 7.5 \\ (+0.2)} &
  \textbf{\shortstack[c]{73.0 $\pm$ 7.0 \\ (+2.2)}} \\
 &
 TS2Vec &
 ERM++ &
 DDG &
 LFR (LODO) &
 68.7 $\pm$ 7.1 &
  \shortstack[c]{70.7 $\pm$ 8.2 \\ (+2.0)} &
  \shortstack[c]{69.6 $\pm$ 6.7 \\ (+0.8)} &
  \shortstack[c]{67.5 $\pm$ 8.4 \\ (-1.2)} &
  \shortstack[c]{71.2 $\pm$ 9.0 \\ (+2.5)} &
  \shortstack[c]{69.9 $\pm$ 7.9 \\ (+1.2)} &
  \shortstack[c]{69.6 $\pm$ 6.7 \\ (+0.9)} &
  \textbf{\shortstack[c]{71.6 $\pm$ 8.9 \\ (+2.9)}} \\ \midrule
\multirow{4}{*}{\rotatebox{90}{\textbf{Cross-Position}}} &
 CNN-PFF &
 ERM++ &
 DDG &
 LFR (LODO) &
 48.6 $\pm$ 9.4 &
  \shortstack[c]{48.0 $\pm$ 11.3 \\ (-0.5)} &
  \shortstack[c]{48.4 $\pm$ 7.2 \\ (-0.2)} &
  \shortstack[c]{46.6 $\pm$ 9.5 \\ (-2.0)} &
  \shortstack[c]{53.9 $\pm$ 7.4 \\ (+5.3)} &
  \shortstack[c]{49.8 $\pm$ 11.7 \\ (+1.2)} &
  \shortstack[c]{48.0 $\pm$ 6.6 \\ (-0.6)} &
  \textbf{\shortstack[c]{54.4 $\pm$ 7.0 \\ (+5.8)}} \\
 &
 IMU-TS &
 ERM++ &
 DDG &
 TF-C (ES) &
 43.8 $\pm$ 7.4 &
  \shortstack[c]{44.7 $\pm$ 7.4 \\ (+0.8)} &
  \shortstack[c]{50.3 $\pm$ 6.6 \\ (+6.5)} &
  \shortstack[c]{45.7 $\pm$ 4.4 \\ (+1.8)} &
  \shortstack[c]{51.5 $\pm$ 7.5 \\ (+7.7)} &
  \shortstack[c]{46.6 $\pm$ 4.1 \\ (+2.8)} &
  \shortstack[c]{50.6 $\pm$ 7.7 \\ (+6.7)} &
  \textbf{\shortstack[c]{53.4 $\pm$ 6.2 \\ (+9.6)}} \\
 &
 ResNet &
 SDMix &
 EFD-Mix &
 LFR (LODO) &
 48.7 $\pm$ 6.8 & 
  \shortstack[c]{51.6 $\pm$ 7.7 \\ (+2.9)} & 
  \shortstack[c]{50.2 $\pm$ 8.5 \\ (+1.5)} & 
  \shortstack[c]{51.5 $\pm$ 6.7 \\ (+2.8)} & 
  \shortstack[c]{51.6 $\pm$ 5.6 \\ (+2.9)} & 
  \shortstack[c]{53.9 $\pm$ 7.4 \\ (+5.1)} & 
  \shortstack[c]{50.6 $\pm$ 7.3 \\ (+1.9)} & \
  \textbf{\shortstack[c]{54.9 $\pm$ 4.5 \\ (+6.1)}} \\
 &
 TS2Vec &
 DIFEX &
 EFD-Mix &
 TF-C (ES) &
 49.0 $\pm$ 6.4 & 
  \shortstack[c]{48.3 $\pm$ 8.7 \\ (-0.7)} & 
  \shortstack[c]{52.6 $\pm$ 5.9 \\ (+3.6)} & 
  \shortstack[c]{48.8 $\pm$ 10.3 \\ (-0.2)} & 
  \shortstack[c]{51.8 $\pm$ 11.2 \\ (+2.8)} & 
  \shortstack[c]{45.1 $\pm$ 8.0 \\ (-3.9)} & 
  \shortstack[c]{55.3 $\pm$ 3.0 \\ (+6.2)} & 
  \textbf{\shortstack[c]{58.5 $\pm$ 7.9 \\ (+9.5)}} \\ \bottomrule
\end{tabular}%
}
\end{table*}

The winning pipelines vary across models and scenarios, with full-configuration gains ranging from $+2.1$ to $+9.5$~pp in CD and from $+5.8$ to $+9.6$~pp in CP. 
Architectural modifications often provide the strongest isolated contribution, particularly DDG for IMU-Transformer, which yields $+5.7$~pp in CD and $+6.5$~pp in CP.

More importantly, \textbf{the largest gains frequently emerge only through component interactions}. 
For IMU-Transformer in CD, ERM++ and DDG improve performance by $+2.3$ and $+5.7$~pp individually, but together they reach $+9.5$~pp. 
An even clearer case occurs for CNN-PFF in CP, where all three components are neutral or detrimental in isolation, yet ERM++ combined with DDG reaches $+5.3$~pp and the complete pipeline $+5.8$~pp.

These results reinforce that the effectiveness of a DG component cannot be inferred reliably from its isolated performance. Weak standalone components can become beneficial when combined with compatible objectives, initializations, or architectural modifications.

\subsection{Robustness Across Distribution-Shift Scenarios}
\label{sec:results-robustness-across-shifts}

The preceding analyses show substantial differences between CD and CP. 
We therefore assess whether a common DG pipeline can remain competitive across both shifts by constructing a Pareto frontier over mean CD and CP accuracy for each model. 
A configuration is Pareto-optimal if no other configuration performs at least as well in both scenarios and strictly better in one.

\begin{figure*}[!hptb]
    \centering
    \includegraphics[width=1.0\textwidth]{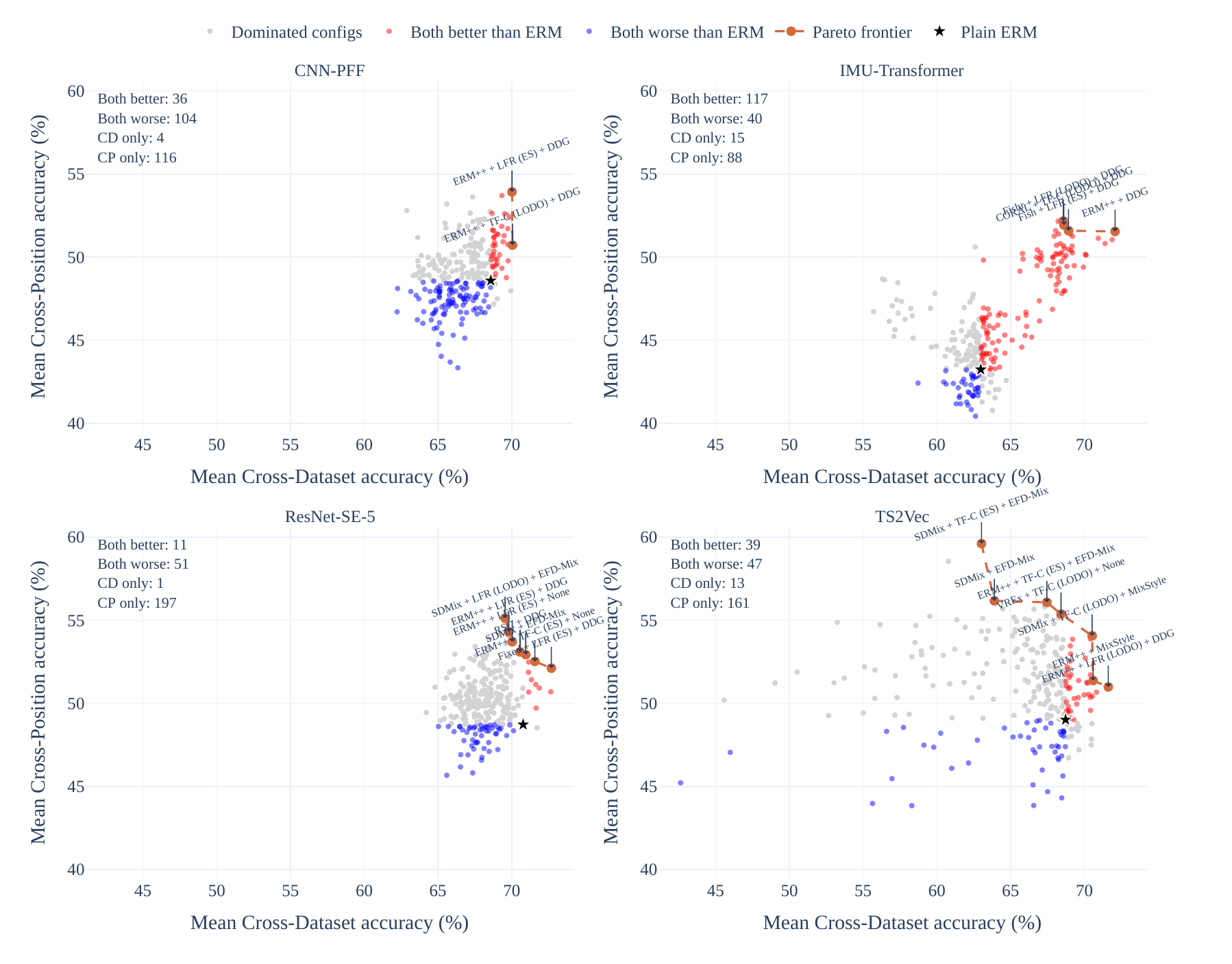}
    \caption{Per-model Pareto analysis across CD and CP. Each point represents an evaluated DG configuration; plain ERM is highlighted with a black star, and the red dashed line connects Pareto-optimal configurations. Dominated configurations are colored relative to ERM: red indicates improvement in both scenarios, blue indicates degradation in both, and gray indicates a trade-off. At the top left, the quadrant counts are shown.}
    \label{fig:pareto_per_model}
\end{figure*}

Figure~\ref{fig:pareto_per_model} shows that cross-shift robustness is strongly model-dependent. 
IMU-Transformer presents the most favorable landscape, with $45\%$ of configurations improving over ERM in both CD and CP, compared with $13.8\%$ for CNN-PFF, $15.0\%$ for TS2Vec, and only $4.2\%$ for ResNet-SE-5. 
The latter exhibits the strongest trade-off, with $75.7\%$ of configurations improving CP while degrading CD.

Across models, configurations that improve CP while degrading CD are far more common than the opposite trade-off, indicating greater headroom for CP improvement. 
IMU-Transformer is the clearest exception, with many configurations providing simultaneous gains across both shifts.

Table~\ref{tab:pareto_per_model} summarizes the Pareto-optimal configurations. IMU-Transformer with ERM++ and DDG yields the strongest balanced result, improving CD by $+9.1$~pp and CP by $+8.3$~pp. DDG also dominates the CNN-PFF frontier and appears in the strongest balanced ResNet-SE-5 configuration, whereas TS2Vec supports a more diverse set of component combinations and exhibits a stronger CD--CP trade-off.

Overall, robust DG across both shifts is possible, but simultaneous improvement depends strongly on the model and the compatibility of the selected DG components.

\begin{table*}[!hptb]
\centering
\caption{Pareto-optimal configurations across CD and CP for each model. $\Delta$CD and $\Delta$CP denote improvements over plain ERM in percentage points (pp), and Mean is the arithmetic mean of CD and CP accuracy. Bold indicates the highest Mean value within each model.}
\label{tab:pareto_per_model}
\small
\begin{tabular}{llllrrrrr}
\toprule
\textbf{Model} &
\textbf{DG Obj.} &
\textbf{Initialization} &
\textbf{Modifier} &
\textbf{CD} &
$\Delta$\textbf{CD} &
\textbf{CP} &
$\Delta$\textbf{CP} &
\textbf{Mean} \\
\midrule

\multirow{2}{*}{CNN-PFF}
& ERM++ & LFR (ES) & DDG & 70.0 & +1.4 & 53.9 & +5.3 & 62.0 \\
& ERM++ & TF-C (LODO) & DDG & 70.1 & +1.5 & 50.7 & +2.1 & 60.4 \\
\midrule

\multirow{4}{*}{IMU-Transformer}
& ERM++ & Random & DDG & 72.1 & +9.1 & 51.5 & +8.3 & 61.8 \\
& Fishr & LFR (LODO) & DDG & 68.6 & +5.6 & 52.2 & +8.9 & 60.4 \\
& CORAL & TF-C (LODO) & DDG & 68.6 & +5.6 & 51.9 & +8.7 & 60.3 \\
& Fish & LFR (ES) & DDG & 68.9 & +5.9 & 51.6 & +8.4 & 60.3 \\
\midrule

\multirow{7}{*}{ResNet-SE-5}
& Fixed & LFR (ES) & DDG & 72.7 & +1.9 & 52.1 & +3.4 & 62.4 \\
& SDMix & LFR (LODO) & EFD-Mix & 69.5 & -1.2 & 55.1 & +6.4 & 62.3 \\
& ERM++ & LFR (ES) & DDG & 69.8 & -1.0 & 54.3 & +5.6 & 62.1 \\
& ERM++ & TF-C (ES) & -- & 71.6 & +0.8 & 52.5 & +3.8 & 62.0 \\
& SDMix & Random & EFD-Mix & 71.0 & +0.2 & 52.9 & +4.2 & 61.9 \\
& ERM++ & LFR (ES) & -- & 70.0 & -0.7 & 53.7 & +5.0 & 61.9 \\
& RSC & Random & DDG & 70.6 & -0.2 & 53.1 & +4.4 & 61.8 \\
\midrule

\multirow{7}{*}{TS2Vec}
& SDMix & TF-C (LODO) & MixStyle & 70.5 & +1.8 & 54.1 & +5.0 & 62.3 \\
& VREx & TF-C (LODO) & -- & 68.4 & -0.3 & 55.4 & +6.4 & 61.9 \\
& ERM++ & TF-C (ES) & EFD-Mix & 67.5 & -1.3 & 56.1 & +7.1 & 61.8 \\
& SDMix & TF-C (ES) & EFD-Mix & 63.0 & -5.7 & 59.6 & +10.6 & 61.3 \\
& ERM++ & LFR (LODO) & DDG & 71.6 & +2.9 & 51.0 & +2.0 & 61.3 \\
& ERM++ & Random & MixStyle & 70.6 & +1.9 & 51.4 & +2.4 & 61.0 \\
& SDMix & Random & EFD-Mix & 63.9 & -4.8 & 56.2 & +7.2 & 60.0 \\
\bottomrule
\end{tabular}
\end{table*}


\subsection{How Winning Configurations Improve Performance}
\label{sec:results-how-winning-improve}

The preceding analyses establish which DG configurations generalize best, but not how their gains emerge at the class level.
We next examine how the winning DG configurations improve predictions at the class level. 
For each model, scenario, and target domain, we compare the winning configuration with plain ERM, yielding $24$ comparisons in CD and $16$ in CP. 
For this post-hoc analysis, we use the seed with the highest target-domain accuracy for each configuration.
We use this choice only for interpretation; it does not affect the source-validation-based selection protocol or previously reported aggregate results.

\subsubsection{Latent-Space Illustration}
\label{sec:results-latent-illustration}

\begin{figure*}[!ht]
    \centering

    \begin{subfigure}[t]{0.49\textwidth}
        \centering
        \includegraphics[width=\linewidth]{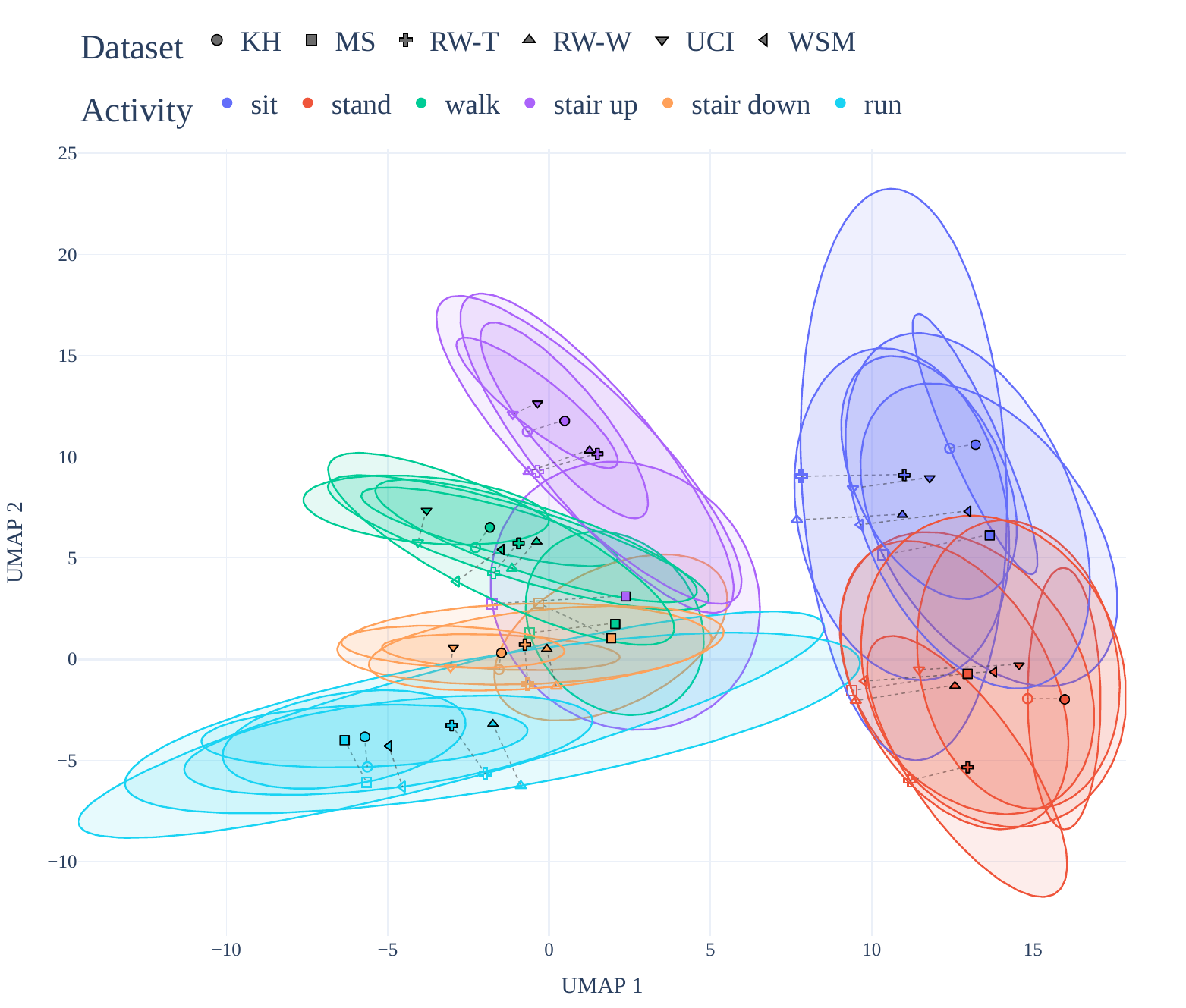}
        \caption{Plain ERM.}
        \label{fig:centroid-visualization-imuts-MS-plain-erm}
    \end{subfigure}
    \hfill
    \begin{subfigure}[t]{0.49\textwidth}
        \centering
        \includegraphics[width=\linewidth]{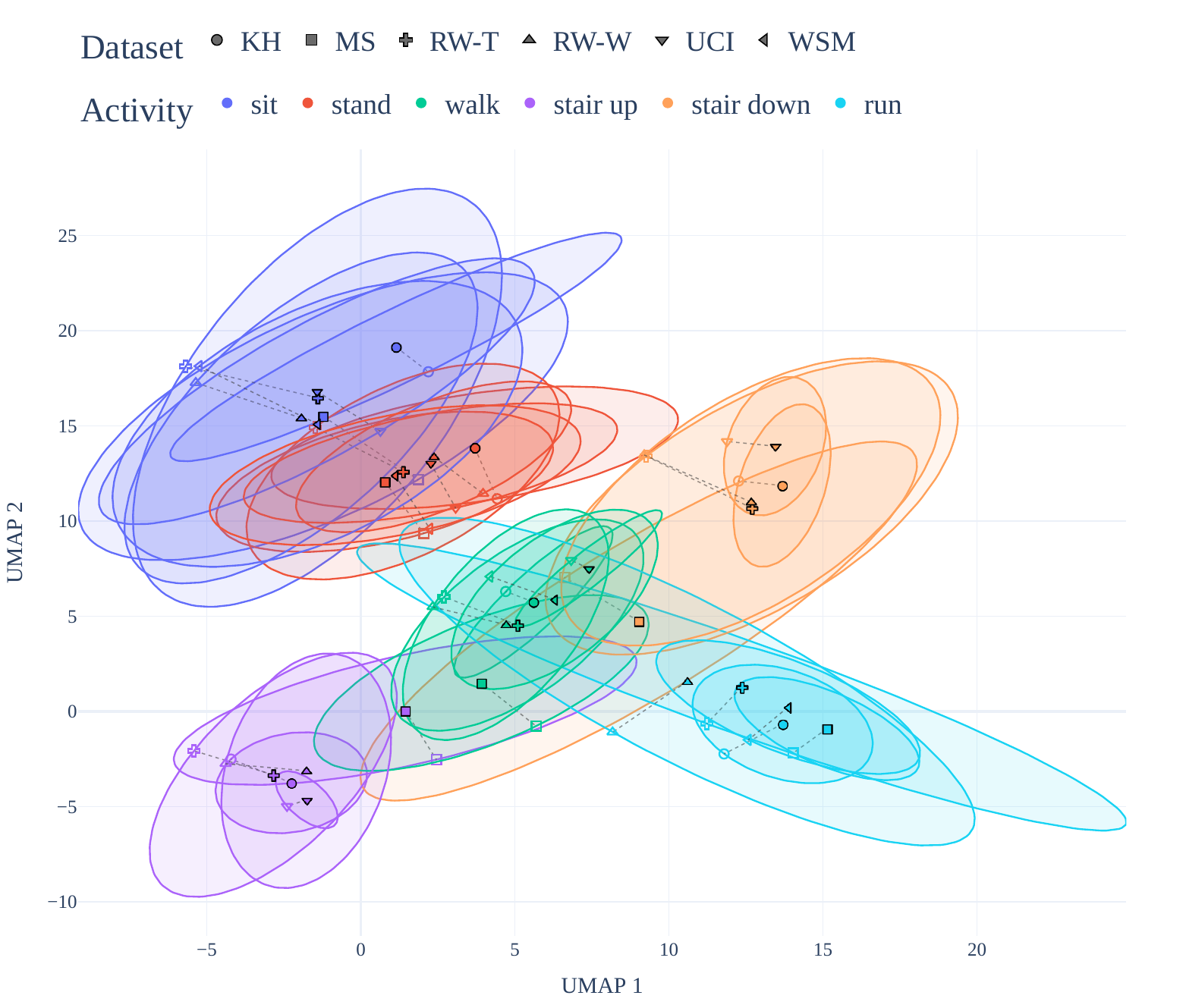}
        \caption{Winning configuration (ERM++ and DDG).}
        \label{fig:centroid-visualization-imuts-MS-best}
    \end{subfigure}

    \caption{Latent-space activity centroids for IMU-Transformer with MS as the held-out CD target. Marker shapes denote datasets and colors denote activities. The winning configuration improves target accuracy by $13.3$~pp and yields clearer separation among difficult locomotion activities, especially for walk and stair-related activities.}
    \label{fig:centroid-visualization-imuts-MS-best-vs-erm}
\end{figure*}

Figure~\ref{fig:centroid-visualization-imuts-MS-best-vs-erm} illustrates IMU-Transformer with MS as the held-out CD target. 
The winning configuration, combining ERM++ and DDG, improves target accuracy by $13.3$~pp, from $64.58\%$ to $77.91\%$. 
Under ERM, several target activity centroids remain separated from their corresponding source-domain clusters, with particularly strong overlap among \textit{walk} and stair activities. 
The winning configuration produces clearer separation among these activities and moves several target centroids toward the corresponding source regions.

This example suggests that improved representation structure contributes to generalization, but does not imply that successful DG requires global source--target alignment. We therefore analyze prediction changes directly.

\subsubsection{Class-Level Error Decomposition}
\label{sec:results-class-error-decomposition}

\begin{figure*}[!hptb]
    \centering
    \includegraphics[width=0.85\linewidth]{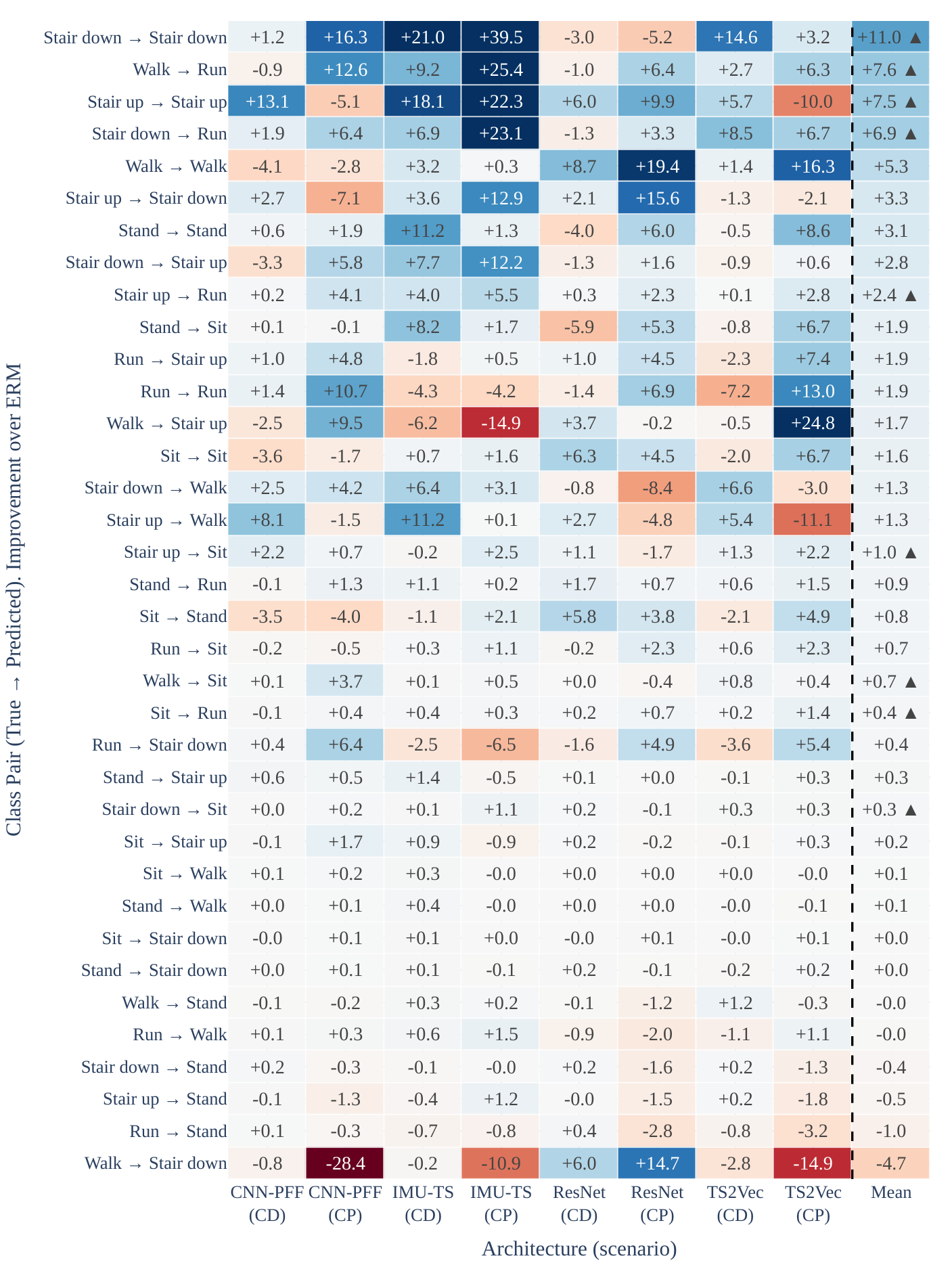}
    \caption{
 Class-level improvement of the winning DG configurations over plain ERM. Each cell reports the change, in percentage points, relative to ERM for a true--predicted class pair. For diagonal entries, positive values denote higher recall; for off-diagonal entries, positive values denote reduced misclassification. Each architecture--scenario column reports the mean change across that scenario's target domains. The ``Mean'' column averages the eight architecture--scenario means, giving equal weight to CD and CP. We compute statistical significance from the $40$ architecture--target differences using the Wilcoxon signed-rank test, with Benjamini--Hochberg correction across class pairs. Upward and downward triangles denote significant positive and negative differences relative to ERM, respectively.
 }
    \label{fig:transitions}
\end{figure*}

Figure~\ref{fig:transitions} shows the difference between the confusion matrices of the winning DG configurations and plain ERM, revealing how DG changes class-level predictions. Positive values consistently indicate improvement over ERM. Diagonal entries ($X\rightarrow X$) denote increased recall of the corresponding class, whereas off-diagonal entries ($X\rightarrow Y$, $X\neq Y$) denote reductions in specific misclassifications. Because each confusion-matrix row is normalized, an increase in diagonal recall corresponds to an equivalent total reduction across the off-diagonal entries of that class.

The strongest significant gains occur for correct recognition of \textit{stair down} ($+11.0$~pp) and \textit{stair up} ($+7.5$~pp), together with reductions in \textit{walk}$\rightarrow$\textit{run} ($+7.6$~pp) and \textit{stair down}$\rightarrow$\textit{run} ($+6.9$~pp) confusions. These results indicate that the winning DG configurations primarily improve discrimination among difficult locomotion classes.

The improvements are not uniformly distributed across all activities. Changes involving \textit{sit} and \textit{stand} are generally smaller, whereas walking, running, and stair activities show larger redistributions. This suggests that DG is especially beneficial for classes with more similar motion patterns, where domain shifts can more strongly alter decision boundaries. At the same time, some class pairs worsen, indicating that the winning configurations redistribute errors rather than reducing every misclassification simultaneously.

Overall, \textbf{winning DG configurations improve generalization through corrections to a limited set of difficult, shift-sensitive decision boundaries}, particularly those involving locomotion and stair-related activities (also illustrated in Figure~\ref{fig:centroid-visualization-imuts-MS-best-vs-erm}).


\subsection{Checkpoint-Selection Headroom Under Domain Shift}
\label{sec:results-headroom}

The preceding analyses show that combining DG components can improve generalization, but deployment also depends on selecting the right checkpoint using only source-domain information. 
We next quantify how much target performance source-validation-based checkpoint selection misses. 
For each training run, we compare the deployable checkpoint selected by mean source-domain validation accuracy (Section~\ref{sec:training_selection}) with the oracle checkpoint that maximizes target-domain accuracy along the same 10,000-step trajectory. 
Because the model, initialization, objective, and optimization path are unchanged, the resulting gap measures checkpoint-selection headroom rather than representational capacity.

\begin{figure*}[!hptb]
    \centering
    \includegraphics[width=1.0\linewidth]{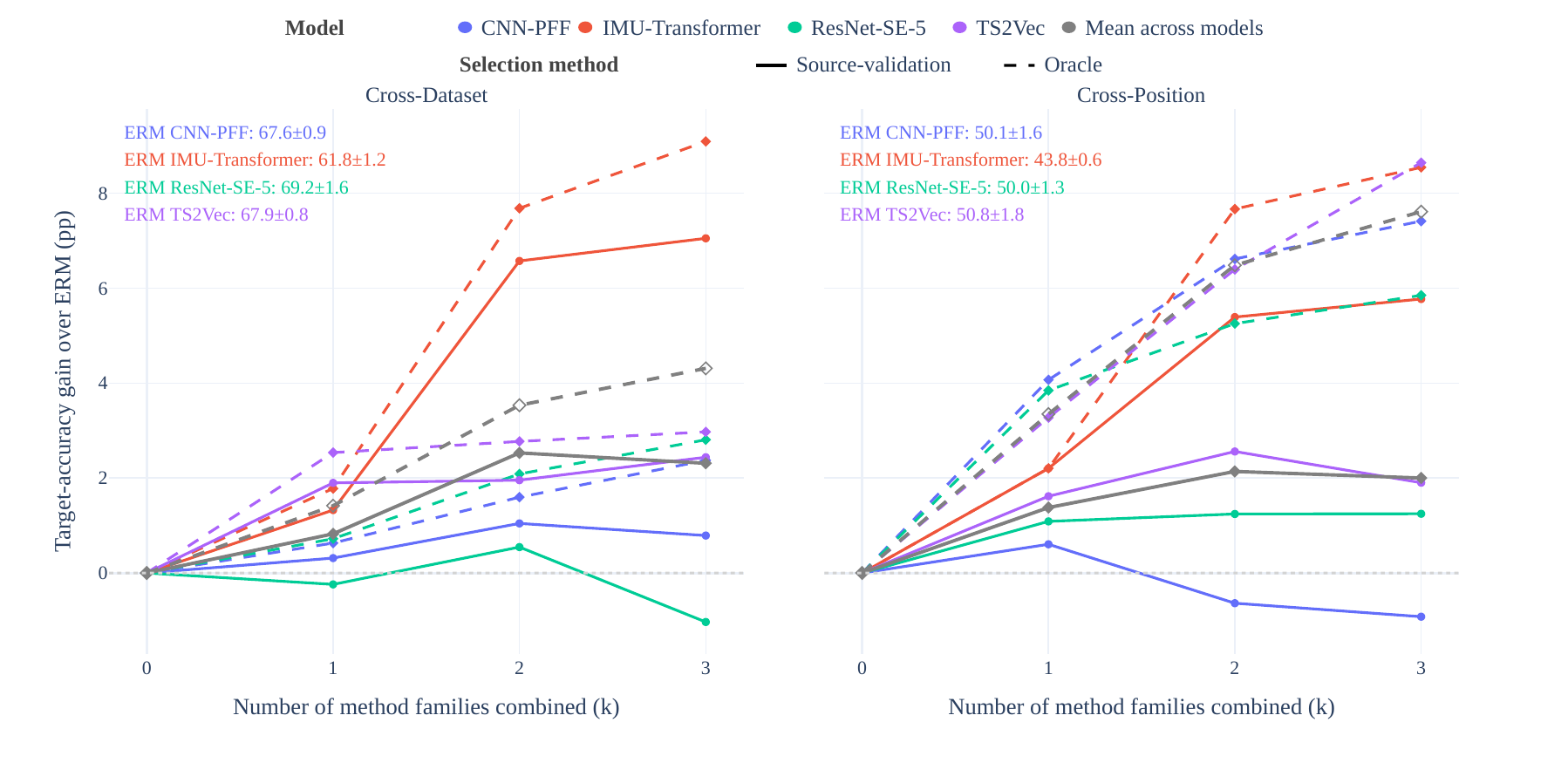}
    \caption{
 Source-validation-selected versus oracle performance across DG complexity levels. 
 For each model, scenario, and number $k$ of active DG dimensions, the figure reports the mean improvement over plain ERM for the same top-10 source-validation-selected configurations used in Section~\ref{sec:results-complementarity}. 
 Solid lines use deployable source-validation-selected checkpoints, while dashed lines use target-domain oracle checkpoints from the same training trajectories.
        \label{fig:oracle_ladder}
 }
\end{figure*}

Figure~\ref{fig:oracle_ladder} compares source-selected and oracle performance for the same top-10 configurations used in Section~\ref{sec:results-complementarity}. 
In CD, source-selected gains are $+0.82$, $+2.53$, and $+2.31$~pp for $k=1,2,3$, whereas oracle gains increase monotonically to $+1.42$, $+3.53$, and $+4.31$~pp. 
The divergence is even larger in CP: source-selected gains reach $+1.38$, $+2.14$, and $+2.00$~pp, compared with oracle gains of $+3.35$, $+6.48$, and $+7.61$~pp.

The selection gap is strongly model-dependent. 
IMU-Transformer retains the largest gains
under both criteria, reaching $+7.05$ versus $+9.09$~pp in CD and $+5.77$ versus $+8.54$~pp in CP at $k=3$. 
More remarkably, ResNet-SE-5 remains below ERM in CD under source selection ($-1.03$~pp) but reaches $+2.80$~pp with oracle selection, showing that target-generalizing checkpoints can exist along trajectories that appear unfavorable under source validation.

At $k=3$, source-validation selection recovers only about $53\%$ of the available oracle gain in CD and $26\%$ in CP. 
Moreover, oracle performance improves monotonically as DG dimensions are added, whereas source-selected performance saturates or declines. 
Thus, \textbf{additional DG generalization potential is often present during training but remains inaccessible under current source-only checkpoint-selection criteria}, making model selection a major bottleneck.

\section{Discussion}
\label{sec:discussion}


\paragraph{Individual DG Components Provide Limited Gains.}
The individual-component analysis in Section~\ref{sec:results-individual} reinforces a recurring finding in DG: strong ERM baselines are difficult to outperform consistently when interventions are evaluated in isolation~\cite{lu2025harood,da2026benchmarking,teterwak2025erm++}. 
ERM++ is the most competitive alternative objective, TF-C (LODO) the most reliable initialization, and DDG the strongest architectural modification, but all gains remain conditional on the model and shift. 
This addresses \textbf{RQ1}.

\paragraph{Component Compatibility Is Essential.}
Addressing \textbf{RQ2}, Sections~\ref{sec:results-complementarity} and~\ref{sec:results-winning-component-attribution} show that the strongest gains emerge from complete DG pipelines rather than isolated techniques. 
Architectural modifications, particularly DDG, provide the clearest direct improvements, but standalone performance often poorly predicts joint behavior. 
In fact, neutral or detrimental components can become beneficial when combined, with five of eight model--scenario pairs exhibiting empirical superadditivity. 
On the other hand, additional dimensions can also introduce redundancy or interference. DG should therefore be treated as a multi-component pipeline-design problem rather than a single-method selection.

\paragraph{DG Gains Depend on Model and Shift.}
Addressing \textbf{RQ3}, Section~\ref{sec:results-robustness-across-shifts} shows substantial model- and shift-dependence. 
IMU-Transformer benefits strongly from DDG, whereas other models exhibit weaker or conflicting interactions. 
CP also presents stronger domain separability than CD. 
In fact, Section~\ref{sec:potential-sources-of-shift} identifies sensor placement as the strongest source of separability in our benchmark. 
Thus, although CD combines more heterogeneous factors, CP systematically varies a particularly influential one, helping explain both its difficulty and its larger headroom for suitable DG configurations.

\paragraph{Improvements Concentrate on Specific Decision Boundaries.}
Section~\ref{sec:results-class-error-decomposition} shows that the largest gains occur on difficult, shift-sensitive boundaries, especially among walking, running, and stair-related activities. 
This suggests that effective DG may depend on identifying and correcting specific behavioral patterns rather than uniformly transforming the entire decision space.

\paragraph{Model Selection Is a Major Bottleneck.}
Section~\ref{sec:results-headroom} shows that substantial target-generalizing performance is reached during training but missed by source-domain selection criteria.
Oracle performance improves monotonically as DG dimensions are added, whereas source-selected performance can saturate or decline. 
At three active dimensions, source validation recovers only $53\%$ of the oracle gain in CD and $26\%$ in CP. 
This suggests that source-domain criteria can favor checkpoints that do not correspond to the best unseen-domain generalization, making source-only model selection a major bottleneck. 
Improving selection criteria may therefore be as important as developing stronger DG objectives or architectures. 
This addresses \textbf{RQ4}.

\subsection{Limitations and Future Work}
\label{sec:limitations}

Despite its scale and controlled design, the benchmark covers a finite set of four model architectures, twelve objective-based alternatives to ERM, three architectural modifications, five initialization strategies, and two shift scenarios. Larger foundation models, generative approaches, and other sensing modalities remain outside the current scope.

The study also focuses exclusively on training-time DG. 
Test-time, continual, and online adaptation can exploit unlabeled target data after deployment and may achieve additional performance gains. 
Evaluating these paradigms under the same datasets, models, and protocol is an important direction for future work.

For scalability and reproducibility, we use domain-balanced sampling and a fixed source-validation split rather than repeated resampling or leave-one-source-domain-out validation~\cite{gulrajanisearch}. 
This preserves subject independence and consistent comparisons across configurations, but future work could investigate source weighting, adaptive sampling, and more robust source-only selection criteria.

Finally, the benchmark is restricted to smartphone-based inertial HAR, limiting external validity, and the class-level and latent-space analyses provide only partial explanations of why specific configurations succeed. Similar controlled studies are needed in wearable, multimodal, medical, and other sensing settings. At the same time, explainable AI and representation-analysis techniques could further identify the temporal, spectral, or local features responsible for cross-domain improvements and failures.

\section{Conclusions}
\label{sec:conclusion}

This work presented a large-scale controlled study of domain generalization for smartphone-based HAR, jointly evaluating representation initialization, architectural modifications, and objective-based DG methods across multiple models and cross-dataset (CD) and cross-position (CP) shifts.

No individual DG component is universally superior. ERM remains a strong objective baseline, TF-C (LODO) is the most reliable initialization strategy evaluated, and DDG provides the clearest architectural gains. 
More importantly, isolated performance poorly predicts joint behavior: compatible components can produce complementary or super-additive gains, while others introduce redundancy or interference.

These effects remain strongly model- and shift-dependent. 
DDG is particularly effective with IMU-Transformer, while CP exhibits stronger domain separability despite its more controlled setting, consistent with sensor placement being a major source of shift. 
The largest class-level gains are also concentrated on difficult locomotion and stair-related decision boundaries.

Source-only model selection further limits realized performance. 
At the highest compositional level, source-validation selection recovers only $53\%$ of the available oracle gain in CD and $26\%$ in CP, even though oracle performance increases as additional DG dimensions are combined.

Overall, effective DG for smartphone-based HAR depends on jointly designing and selecting compatible combinations of model architecture, representation initialization, training objective, and model-selection strategy. 
Future DG methods should therefore consider these components jointly rather than relying on any one of them in isolation.

\section*{Acknowledgments}

We used AI tools to improve the readability of this paper. In particular, we used ChatGPT recommendations to edit all sections of the manuscript for grammar, clarity, and readability. We also used Grammarly to check the manuscript's grammar and spelling.

This work was supported in part by the Ministry of Science, Technology, and Innovation of Brazil (MCTI), with resources granted by Federal Law 8.248 of October 23, 1991, through the Priority National Innovation Program administered by Softex (PPI-Softex) under Grant 01245.003479/2024-10; in part by the Coordination for the Improvement of Higher Education Personnel-Brazil (CAPES) under Grant 88887.999360/2024-00; in part by Brazilian National Research Council (CNPq) under Grant 315399/2023-6 and Grant 302458/2022-0; and in part by the São Paulo Research Foundation (FAPESP) under Grant 2013/08293-7 and Grant 2023/12865-8.

\bibliographystyle{ieeetr}
\bibliography{bibliography}

\newpage

\appendix

\section{Training Procedure and Hyperparameters}
\label{sec:appendix-models-hyperparameters}

This section provides additional details on the training procedure and hyperparameter search grids used in our evaluation.

\subsection{Number of Training Steps}
\label{sec:appendix-training_steps-ablation}

All supervised configurations are trained for a fixed budget of 10,000 optimization steps. 
We assess whether this optimization budget is sufficient by analyzing the training step at which the selected checkpoint is obtained. 
If checkpoints are consistently selected near the end of training, the optimization budget may still limit performance; earlier selections suggest that additional training is unlikely to provide consistent gains.

\begin{figure*}[!hptb]
    \centering
    \includegraphics[width=1.0\textwidth]{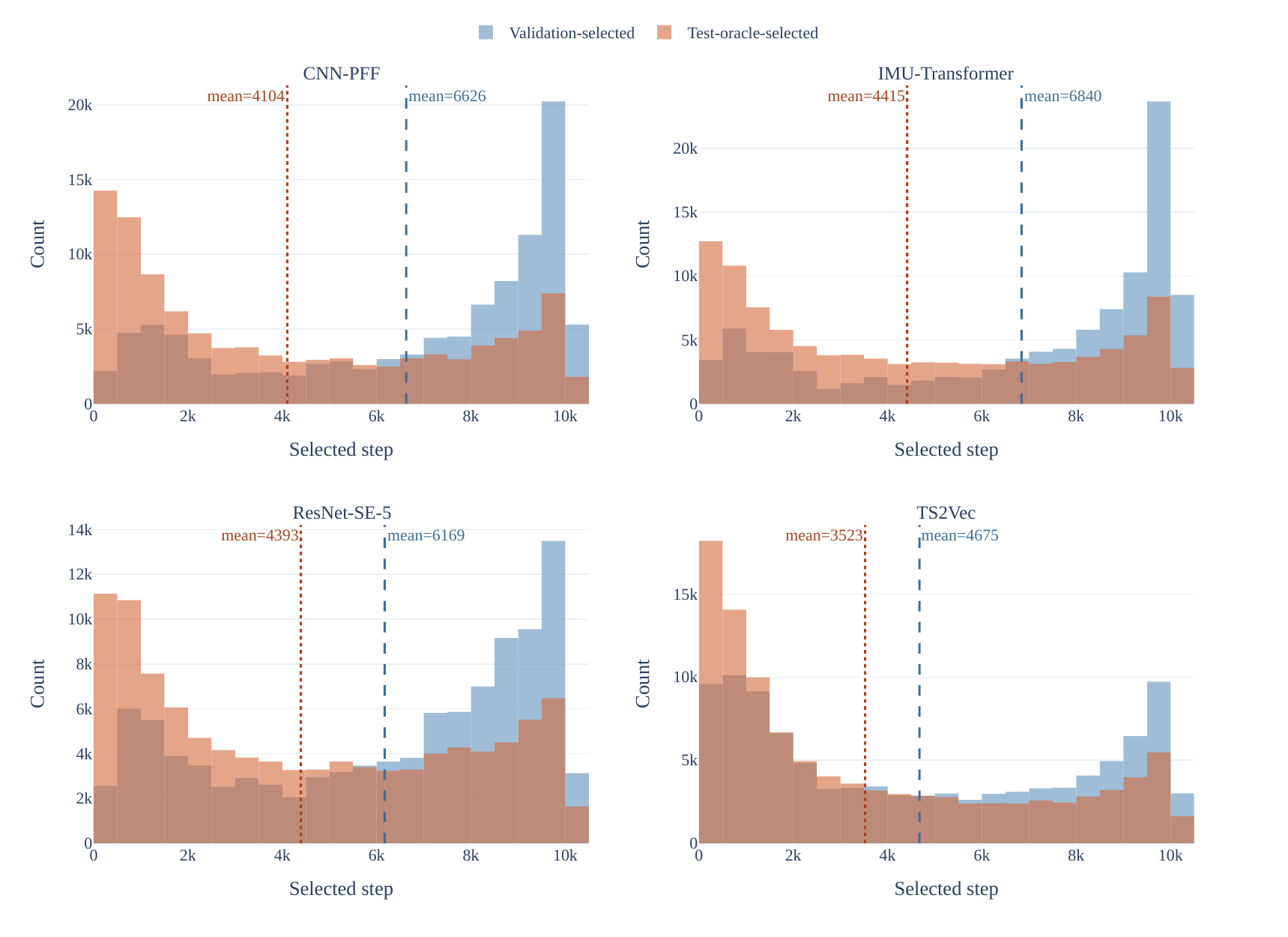}
    \caption{Distribution of the selected training step for each backbone under source-validation-based and oracle target-domain checkpoint selection. The source-validation protocol corresponds to the deployable procedure used throughout the benchmark, whereas the oracle protocol selects the checkpoint with the highest target-domain accuracy.}
    \label{fig:convergence_histograms}
\end{figure*}

Figure~\ref{fig:convergence_histograms} shows the selected-step distributions under two checkpoint-selection protocols. The first is the deployable source-validation protocol used throughout the benchmark, which selects the checkpoint with the highest mean validation accuracy across the source domains. 
The second is an oracle protocol that selects the checkpoint with the highest target-domain accuracy, the upper-bound performance achievable with perfect knowledge of the target domain during training.

Under source validation, checkpoints are generally selected during the middle and later stages of training. 
The mean selected step is approximately 6,600 for CNN-PFF, 6,800 for IMU-Transformer, 6,200 for ResNet-SE-5, and 4,700 for TS2Vec. 
Moreover, for CNN-PFF, IMU-Transformer, and ResNet-SE-5, at least 25\% of the runs select checkpoints after roughly 8,300 steps, indicating continued improvement on the source domains. 
TS2Vec is consistently selected earlier and is therefore the least likely backbone to benefit from a larger optimization budget.

The oracle protocol consistently selects earlier checkpoints. 
For example, TS2Vec shifts from a mean of approximately 4,700 steps under source validation to 3,500 under oracle selection, with similar trends for the other backbones. 
This indicates that source-domain validation and target-domain generalization typically peak at different stages of training: source validation often favors later checkpoints even after target-domain performance has begun to decline or stagnate.

Thus, the 10,000-step budget is sufficient to capture the optimization behavior of all evaluated backbones. 
Most oracle-optimal checkpoints occur well before the training limit, suggesting that extending training would primarily improve source-domain fitting rather than generalization to unseen domains. 
Although some source-validation checkpoints are still selected near the end of training, the remaining generalization gap is more strongly tied to checkpoint selection than to insufficient optimization.

\subsection{Hyperparameter Search Grids}
\label{sec:appendix-hparam-grids}

Most evaluated DG methods introduce hyperparameters controlling optimization, regularization, or feature transformation. 
To support a fair comparison, we define method-specific search grids using values recommended in the original publications, configurations adopted by DomainBed~\cite{gulrajanisearch} when applicable, and additional values identified through preliminary experiments.

Tables~\ref{tab:dg_hparam_grid} and~\ref{tab:arch_modifier_grid} summarize the search spaces for objective-based DG methods and architectural modifications, respectively. 
For a given objective-based method, its candidate configurations are formed from the Cartesian product of the hyperparameter values listed for that method. 
We apply the same procedure independently to each architectural modification. 
Method-specific parameters are therefore varied only within the method to which they belong; hyperparameters from different objective-based methods are not crossed.

We use Adam with two learning rates, $10^{-3}$ and $10^{-4}$, applied uniformly across all methods, resulting in one run per learning rate.
Since some methods yield substantially larger grids than others, up to five configurations are uniformly sampled from each grid for each learning rate\footnote{Default and best hyperparameter configurations recommended by the original authors are always included and take precedence over randomly sampled configurations.}. When a method has fewer than five possible configurations, we evaluate all configurations. This strategy balances computational cost while ensuring comparable hyperparameter exploration across methods.

The broader benchmark follows a full-factorial design across DG component families. We evaluate each sampled objective-based configuration with every initialization strategy and with both the original backbone and each architectural-modification configuration. This design enables analysis of the effects of the objective, initialization, and architecture, both independently and jointly.

We train every configuration independently using three random seeds. Within each run, we select checkpoints using mean source-domain validation accuracy, following Section~\ref{sec:training_selection}.

We then select hyperparameters as required by each analysis. 
For each candidate configuration within an experimental setting, we average source-validation accuracy across the three seeds. 
We select the configuration with the highest mean source-validation accuracy and report its target-domain performance as the mean and standard deviation across the corresponding three runs. 
We never use target-domain labels for checkpoint or hyperparameter selection.

This procedure evaluates every method-component combination under its best source-validation-selected configuration. 
It thereby reduces confounding from clearly suboptimal hyperparameters while preserving the practical DG constraint that all model-development decisions must be made without access to the held-out target domain.

\begin{table*}[!hptb]
\centering

\caption{Hyperparameter search grids for the objective-based domain generalization techniques. For each technique, this grid is also crossed with an outer Adam learning-rate grid of $\{10^{-3}, 10^{-4}\}$, applied uniformly across all techniques.
}
\label{tab:dg_hparam_grid}
\resizebox{\textwidth}{!}{
\begin{tabular}{@{}l l p{7.6cm} p{3.2cm}@{}}
\toprule
\textbf{Technique} & \textbf{Parameter} & \textbf{Description} & \textbf{Values explored} \\
\midrule

CORAL
& \texttt{mmd\_gamma} & CORAL/MMD feature-covariance alignment penalty weight & $10^{\mathcal{U}(-1,1)}$ \\
\midrule

\multirow{4}{*}{DIFEX}
& \texttt{alpha} & Weight of the Fourier-domain-invariant feature loss & $\{0.001\mcomma 0.01\mcomma 0.1\mcomma 0.5\mcomma 1.0\mcomma 10.0\}$ \\
& \texttt{beta} & Weight of the cross-domain feature alignment loss & $\{0.01\mcomma 0.1\mcomma 0.5\mcomma 1.0\mcomma 10.0\}$ \\
& \texttt{lam} & Weight of the teacher-distillation loss & $\{0.01\mcomma 0.1\mcomma 0.5\mcomma 1.0\}$ \\
& \texttt{disttype} & Distance metric for Fourier feature alignment & $\{$norm-1-norm, 2-norm, cos$\}$ \\
\midrule

ERM & --- & Baseline. No hyperparameters searched & N/A \\
\midrule

ERM++
& \texttt{linear\_steps} & Number of linear-probing steps before full fine-tuning & $\{0, 100\}$ \\
\midrule

Fish
& \texttt{meta\_lr} & Meta-learning rate of the inter-domain gradient step & $\{0.05, 0.1, 0.5\}$ \\
\midrule

\multirow{3}{*}{FishR}
& \texttt{fishr\_lambda} & Weight of the gradient-variance (Fishr) penalty & $10^{\mathcal{U}(1,4)}$ \\
& \texttt{penalty\_anneal\_iters} & Training steps before the Fishr penalty is annealed in & $\mathcal{U}(0, 5000)$ \\
& \texttt{ema} & Exponential moving average decay of the gradient covariance & $\mathcal{U}(0.90, 0.99)$ \\
\midrule

\multirow{4}{*}{FIXED}
& \texttt{mixup\_alpha} & Beta-distribution shape parameter for mixup interpolation & $\{0.1\mcomma 0.2\mcomma 0.5\mcomma 1.0\mcomma 10.0\}$ \\
& \texttt{grl\_alpha} & Gradient-reversal-layer weight of the domain-adversarial branch & $\{0.01\mcomma 0.1\mcomma 0.5\mcomma 1.0\}$ \\
& \texttt{margin\_gamma} & Scale of the large-margin softmax penalty & $\{10\mcomma 10^{2}\mcomma 10^{3}\mcomma 10^{4}\mcomma 10^{5}\}$ \\
& \texttt{margin\_top\_k} & Number of top classes considered by the margin penalty & $\{1, 2, 5\}$ \\
\midrule

GroupDRO
& \texttt{groupdro\_eta} & Step size of the robust group-weight update & $10^{\mathcal{U}(-3,-1)}$ \\
\midrule

\multirow{5}{*}{LAG}
& \texttt{bottleneck\_dim} & Dimensionality of the bottleneck layer before alignment & $\{128, 256, 512\}$ \\
& \texttt{layer} & Backbone representation aligned (original vs.\ batch-normalized) & $\{$ori, bn$\}$ \\
& \texttt{classifier} & Classifier head type (linear vs.\ weight-normalized) & $\{$linear, wn$\}$ \\
& \texttt{mmd\_gamma} & MMD feature alignment penalty weight & $10^{\mathcal{U}(-1,1)}$ \\
& \texttt{rela\_gamma} & Relation-alignment penalty weight & $10^{\mathcal{U}(-1,1)}$ \\
\midrule

\multirow{3}{*}{RDM}
& \texttt{rdm\_lambda} & Weight of the risk-distribution-matching penalty & $\mathcal{U}(0.1, 10.0)$ \\
& \texttt{rdm\_penalty\_anneal\_iters} & Training steps before the RDM penalty is annealed in & $\mathcal{U}(800, 2700)$ \\
& \texttt{variance\_weight} & Weight of the auxiliary variance regularization term & $10^{\mathcal{U}(-5,-1)}$ \\
\midrule

\multirow{2}{*}{RSC}
& \texttt{rsc\_f\_drop\_factor} & Fraction of top-activated features masked (forward pass) & $\mathcal{U}(0.0, 0.5)$ \\
& \texttt{rsc\_b\_drop\_factor} & Fraction of top-gradient features masked (backward pass) & $\mathcal{U}(0.0, 0.5)$ \\
\midrule

\multirow{5}{*}{SDMix}
& \texttt{mixupalpha} & Beta-distribution shape parameter for mixup interpolation & $\{0.1\mcomma 0.2\mcomma 0.5\mcomma 1.0\}$ \\
& \texttt{mixup\_ld\_margin} & Scale of the large-distance mixup margin penalty & $\{10\mcomma 10^{2}\mcomma 10^{3}\mcomma 10^{4}\mcomma 10^{5}\}$ \\
& \texttt{top\_k} & Number of top classes considered by the margin penalty & $\{1, 2, 5\}$ \\
& \texttt{disttype} & Distance metric for the margin penalty & $\{$1-norm, 2-norm, cos$\}$ \\
& \texttt{normstyle} & Feature normalization strategy (max vs.\ average) & $\{$max, avg$\}$ \\
\midrule

\multirow{2}{*}{VREx}
& \texttt{vrex\_lambda} & Weight of the variance-risk-extrapolation penalty & $10^{\mathcal{U}(-1,5)}$ \\
& \texttt{vrex\_penalty\_anneal\_iters} & Training steps before the VREx penalty is annealed in & $10^{\mathcal{U}(0,4)}$ \\

\bottomrule
\end{tabular}
}
\end{table*}

\begin{table*}[!hptb]
\centering
\small
\begin{threeparttable}

\caption{Configurations explored for the architectural modification strategies. Unlike the objective-based techniques, these are not searched via a random grid: each backbone is instantiated once per listed configuration, and MixStyle/EFD-Mix additionally ship two fixed $\alpha$ settings.
}
\label{tab:arch_modifier_grid}

\begin{tabular}{@{}l l p{7.5cm} l@{}}
\toprule
\textbf{Technique} & \textbf{Parameter} & \textbf{Description} & \textbf{Values explored} \\
\midrule

Default & --- & Backbone used as-is, no architectural modification & N/A \\
\midrule

\multirow{3}{*}{MixStyle}
& \texttt{alpha} & Beta-distribution shape controlling feature-statistics mixing strength & $\{0.1, 0.2\}$ \\
& \texttt{p} & Application probability & $p=0.5$ \\
& \texttt{activated\_blocks} & Backbone blocks where the layer is inserted & First three layers of the backbone \\
\midrule

\multirow{3}{*}{EFD-Mix}
& \texttt{alpha} & Beta-distribution shape controlling exact feature-distribution mixing strength & $\{0.1, 0.2\}$ \\
& \texttt{p} & Application probability & $p=0.5$ \\
& \texttt{activated\_blocks} & Backbone blocks where the layer is inserted & First three layers of the backbone \\
\midrule

DDG
& \texttt{activated\_blocks}
& Backbone blocks where the layer is inserted
& First three layers of the backbone\tnote{a} \\
\bottomrule
\end{tabular}

\begin{tablenotes}[flushleft]
\footnotesize
\item[a] For IMUTransformer, DDG is inserted only in the input projection layer.
\end{tablenotes}

\end{threeparttable}
\end{table*}

\subsection{Number of Configurations for Complementarity Analysis}
\label{sec:appendix-number-of-configurations}

The complementarity analysis averages the ten best source-validation-selected configurations rather than relying on a single configuration or the entire search space. 
This choice provides a small high-performing subset that reduces sensitivity to an unusually favorable configuration while avoiding dilution from the large number of weak or small beneficial combinations.

\begin{figure*}[!hptb]
    \centering
    \includegraphics[
 width=0.90\linewidth,
 trim={0 0 0 0},
 clip
 ]{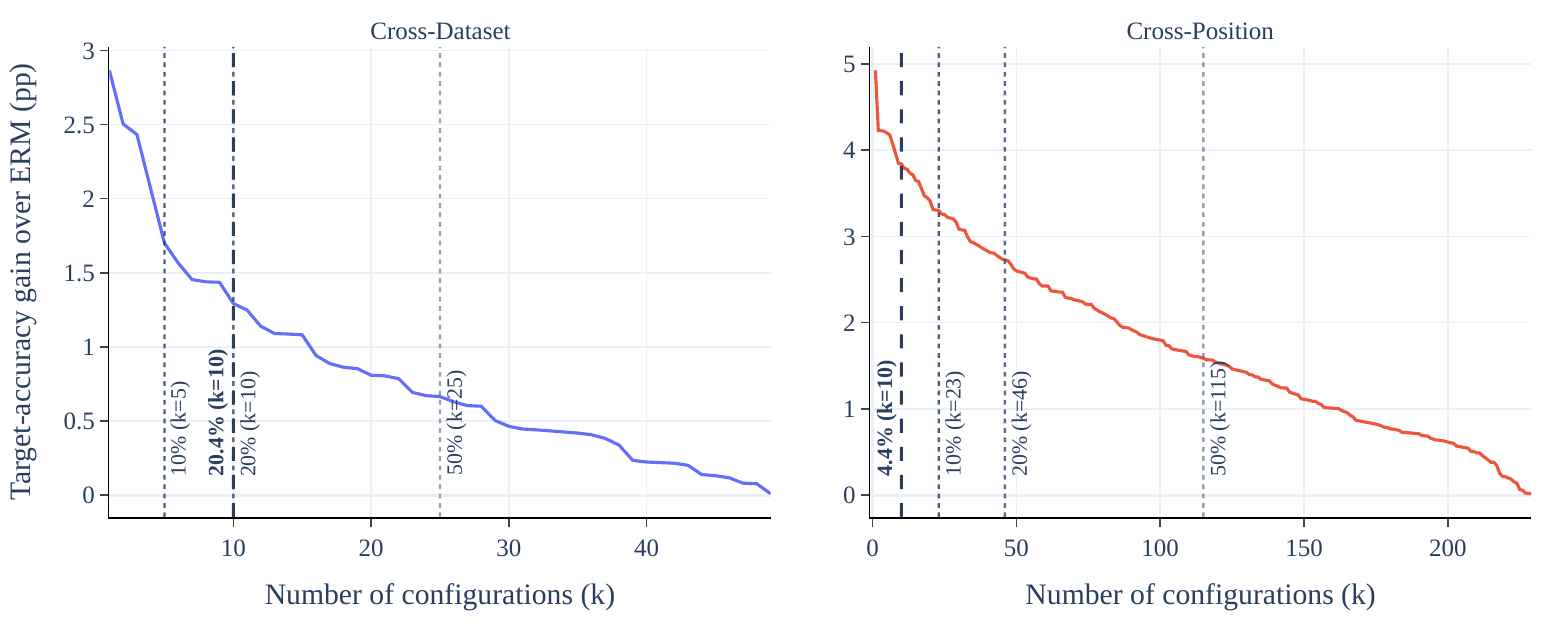}
    \caption{Target-accuracy gain over ERM as a function of the number of positively improving configurations, ranked from highest to lowest gain. The dashed lines indicate the top $10\%$, $20\%$, and $50\%$ of improving configurations and the fixed choice $k=10$ used in the complementarity analysis.}
    \label{fig:number-of-configurations-to-explore}
\end{figure*}

Figure~\ref{fig:number-of-configurations-to-explore} shows the target-accuracy gain over ERM after ranking all configurations with improvement from best to worst. There are $49$ such configurations in CD and $229$ in CP. 
The figure marks the top $10\%$, $20\%$, and $50\%$ of these configurations, together with the fixed choice $k=10$ used in the main analysis.

For CD, $k=10$ corresponds to approximately the top $20.4\%$ of configurations that outperform ERM, while for CP it corresponds to the top $4.4\%$. 
Despite the different search-space sizes, $k=10$ remains concentrated in the high-performing region of both curves and avoids including the long tail of configurations with progressively smaller gains. 
It therefore provides a reasonable compromise between representing multiple competitive configurations and retaining focus on the strongest portion of the search space.

\end{document}

%% file: macros.tex
\usepackage{xspace}
\usepackage[table]{xcolor}
\usepackage{amssymb}
\usepackage{environ}
\usepackage{ifthen}
\usepackage{marginnote}
\usepackage{graphicx}
\usepackage{tikz}
\usepackage{soul}

\newcommand{\italico}[1]{\emph{#1}}

\newcommand{\ie}{\italico{i.e.}\xspace}

\newcommand{\etal}{\italico{et al.}\xspace}
\newcommand{\iid}{\italico{i.i.d.}\xspace}

\newcommand{\ck}{\checkmark}

\DeclareRobustCommand{\circnum}[1]{%
  \tikz[baseline=(char.base)]{%
    \node[
      shape=circle,
      fill=black,
      inner sep=1pt,
      text=white
    ] (char) {\small\bfseries #1};
  }%
}

\newif\iffinalversion
\newif\iffinalcolor

\finalversionfalse
\finalcolorfalse

\iffinalcolor
    \def\borincolor{black}
    \def\otaviocolor{black}
\else
    \def\borincolor{blue}
    \def\otaviocolor{teal}
\fi

\iffinalversion

    \newcommand{\TODO}[1]{}
    \newcommand{\EB}[1]{}

    \newcommand{\EBRM}[1]{}

    \newcommand{\ON}[1]{}

\else

    \newcommand{\TODO}[1]{%
        \textcolor{red}{(TODO: #1)}%
    }

    \newcommand{\EB}[1]{%
        \textbf{\textcolor{\borincolor}{(EB: #1)}}%
    }

    \newcommand{\EBRM}[1]{%
        \textcolor{lightgray}{(#1)}%
    }

    \newcommand{\ON}[1]{%
        \textbf{\textcolor{\otaviocolor}{(ON: #1)}}%
    }

\fi

\newif\ifshowchanges
\newif\ifshowtodos

\showchangestrue
\showtodostrue

\ifshowchanges
    
\else
    
\fi

\ifshowtodos
    
\else
    
\fi